\documentclass[twocolumn]{svjour3}          

\journalname{}
\date{}

\newcommand{\vecteur}[2]{{\boxmaxdepth=1000pt\relax\overrightarrow{#1#2}}}
\usepackage{graphicx}
\usepackage{amsmath}
\usepackage{amssymb}
\usepackage{balance}
\begin{document}
\newtheorem{theor}{Theorem}

\title{Object Pose from 2-D to 3-D Point and Line Correspondences
\thanks{This work has been supported by the Esprit programme through the SECOND project (Esprit-BRA No. 6769).} }
\titlerunning{Object Pose from 2-D to 3-D Point and Line Correspondences}
\author{Thai Quynh Phong \and Radu Horaud \and Adnan Yassine \and Pham Dinh Tao}
\authorrunning{T. Q. Phong, R. Horaud, A. Yassine \& P. D. Tao}
\institute{
Thai Quynh Phong \& Radu Horaud \at
Inria Rh\^{o}ne-Alpes\\
46, avenue F\'{e}lix Viallet\\
38031 Grenoble FRANCE
\and
Adnan Yassine \at
D\'epartement de Math\'ematiques\\
Universit\'e de Nancy I\\
BP 239\\
54506 Vandoeuvre-les-Nancy FRANCE
\and
Pham Dinh Tao \at
D\'epartement G\'enie Math\'ematique\\
INSA -- Rouen\\
BP 08 Place Emile Blondel\\
76131 Mont-Saint-Aignan FRANCE
}
\maketitle

\begin{abstract} In this paper we present a method for 
optimally estimating the rotation and translation between a camera and a 3-D object
from point and/or line correspondences. First we devise an error function and second we show 
how to minimize this error function. The quadratic nature of this function
is made possible by representing rotation and translation with a dual number quaternion. We
provide a detailed account of the computational aspects of a trust-region optimization
method. This method compares favourably with Newton's method which has extensively been used
to solve the problem at hand, with Faugeras-Toscani's linear method
\cite{FaugerasToscani86} for calibrating a camera, and with the Levenberg-Marquardt non-linear
optimization method. 
Finally we present some experimental results which de\-mon\-strate the robustness of our method
with respect to image noise and matching errors.
\end{abstract}

\section{Introduction}
\label{section:Introduction}

The problem of determining the position and orientation of an object with
respect to a camera has many relevant applications in computer vision:
object positioning, camera calibration, hand-eye calibration, docking for land and space
mobile robots, and cartography.
This
problem is also known as the {\em perspective n-point problem}, {\em exterior
(or extrinsic) camera calibration problem}, or {\em camera location and
orientation problem}, and can be stated more formally as follows: Given a set
of points that are described in an {\em object centered frame}, given the
projections of these points onto an image, and given a projection model and
the parameters of this model, determine the rigid transformation (rotation and
translation) between the object centered frame and the camera centered frame.

\subsection{Background}

Previous approaches attempting to solve this problem fall into two categories:
\begin{itemize}
\item {\em Closed-form solutions}. Whenever the number of point correspondences
is limited one may devise a closed-form
solution. Such solutions exist for 3 points (Fischler \& Bolles 
\cite{FischlerBolles81}, 4 coplanar points (Hung \& al. \cite{HungYehHarwood85}), 4 points in
general position (Horaud \& al. \cite{HoraudConioLeboulleuxLacolle89}, Holt \&
Netravali \cite{HoltNetravali91}), and 3 lines in general position (Dhome \& al.
\cite{Dhome89}, Chen \cite{Chen91a}).
\item {\em Numerical solutions}. Whenever the number of point correspondences is
large (greater than 6) closed form solutions are not efficient because one has
to solve for a set of non-linear equations where the number of equations is
larger than the number of unknowns (there are six unknowns, 3 for the rotation
and 3 for the translation). Therefore, iterative numerical solutions are
generally required. Some of these iterative solutions are discussed below.
\end{itemize}

Since the object pose from a single view problem is nonlinear, choices for
(i)~the mathematical representation of the problem, (ii)~the error function to
be minimized, and for (iii)~the optimization method are crucial.

Ganapathy \cite{Ganapathy84bis} and others (see for example
\cite{FischlerBolles81}) mention a linear solution for point correspondences.
Rotation is represented by a 3$\times$3 matrix. The linear method is extremely
susceptible to noise mainly because the orthogonality constraints associated
with the rotation matrix are not taken into account.

Yuan \cite{Yuan89} proposed to separate the rotational component of the problem from
the translational one and he concentrated on the estimation of the rotation
parameters. The rotation is represented by an orthogonal matrix and the
solution is given by the common root of six quadratic equations. The common
root is then found using Newton's iterative gradient method. However the
author noticed that local optima occur when gradient techniques are used.
Several local minima correspond to the nonlinear nature of the problem. The
global minimum can be reached only by properly initializing the iterative
algorithm.

Lowe \cite{Lowe87} used Newton's method as well for estimating the orientation
and location of an object with respect to a camera. The error function to be
minimized sums up the squares of the distances between object projections and
image measurements (this distance is measured in the image plane). As with
Yuan's method, Lowe noticed some problems with Newton's method and in a
subsequent paper he suggested how to deal with the initialisation and
stability problems \cite{Lowe91}. The stabilization method that he suggests works
well, provided that good initial estimates are available for some of the
sought parameters.

Haralick \& al. \cite{Haralicketal89} present the exterior camera calibration
problem in a more general context of pose estimation from corresponding
2-D/2-D, 2-D/3-D, and 3-D/3-D point data. The authors give special attention
to the modelling of image noise, including a noise model to represent
incorrect matches. Moreover, classical least-square estimation is converted
into an iterative re-weighted least square estimation. In the particular case
of 2-D (perspective projection)/3-D pose estimation, Haralick \& al. notice
that a 1958 German dissertation surveyed nearly 80 different solutions from
the photogrammetry literature! Haralick's et al. own solution represent
rotation with the Euler angles and the equations are linearized by Newton's
first order approximation. 

Liu \& al. \cite{LiuHuangFaugeras90} examined alternative iterative approaches to solving for the
viewing parameters. The authors noticed that with line correspondences
rotation can be easily separated from translation and that a line
correspondence can be created from 2 point correspondences. Once the rotation
has been determined, the estimation of translation is reduced to a linear
problem. The rotation is  represented by the Euler angles. The authors
linearize the error function. They noticed that their method worked well only
when the three Euler angles are less than $30^{0}$.

Using the mathematical formulation suggested by Liu \& al. \cite{LiuHuangFaugeras90}, Kumar \&
Hanson \cite{KumarHanson89} examined two minimization methods: an iterative technique that
linearizes the error function and which requires a good initial estimate and a
least median of squares technique. The error function associated with the
latter is not a differentiable function and a combinatorial algorithm is
therefore required. Such a combinatorial approach is able to eliminate
outliers but it is very time consuming because it considers all triplets of
line correspondences. See for instance the paper of Fischler \& Bolles \cite{FischlerBolles81}
who suggested a heuristic to reduce the combinatorial complexity of such
methods.

\subsection{Approach}

In the light of the above discussion a robust and accurate method is still to be proposed.
In this paper we devise a method for solving the object pose problem. The method is
tailored as follows:
\begin{itemize}
\item Section 2 -- each line correspondence (or equivalently each pair of point
correspondences) provides two constraints which express that the object line,
its corresponding image projection, and the center of projection of  the
camera are coplanar. This approach has already been used by Horaud \& al.
\cite{HoraudConioLeboulleuxLacolle89}, 
Dhome \& al. \cite{Dhome89}, Chen \cite{Chen91a}, Liu \& al. \cite{LiuHuangFaugeras90} and Kumar
\& Hanson \cite{KumarHanson89}.

The rigid transformation, whose parameters are the unknowns of the
problem, is represented by a {\em dual number quaternion}. With this
representation the constraints mentioned above become  quadratic equations.
Therefore, each line correspondence provides two quadratic constraints.
In the past, unit quaternions have been used to represent and estimate 
rotation between two sets of 3-D features (points, lines, planes, surface
patches) i.e., Faugeras \& H\'ebert \cite{FaugerasHebert86} and Horn \cite{Horn87-quat}. 
Puget \& Skordas
\cite{PugetSkordas90a} used unit quaternions to estimate rotation in the context of exterior
camera calibration but their method requires that the translation is estimated
first. 
Walker \& al. \cite{WalkerShaoVolz92} introduced dual number quaternions in
computer vision and they suggested an elegant closed-form solution for the
3-D/3-D pose estimation problem from point and vector correspondences. 
Dual number quaternions are a mathematical representation of screws -- a
well known parameterization of the displacement group. 
To our knowledge
there has been no attempt to use dual number quaternions in conjunction with the
exterior camera calibration (2-D/3-D pose) problem.
\item Section 3 -- 
we show that constraints based on point correspondences lead to an error
function that have the same structure as for line correspondences. Indeed,
each point correspondence provides two qua\-dra\-tic constraints. Nevertheless,
with point correspondences, rotation and translation cannot be separated as it
is the case with line correspondences.
\item Section 4 -- a non linear 
numerical optimization tech\-nique is described and used for estimating
the best rigid transformation. The error function to be minimized over the
pose parameters is the sum of squares of the quadratic constraints just
described. Unlike most of previous approaches in computer vision, we use a
local
second order approximation of the error function. More specifically we use a
{\em trust region} optimization method. The idea of using a trust
region goes back to Soresen \cite{kn:So} and Mor\'{e} \cite{kn:Mo83}. 
Trust region (or restricted step) methods are briefly discussed by Fletcher 
\cite{Fletcher90}.
Yassine \cite{kn:Ya} compared
this method with other second-order and first-order optimization methods. The
general idea of the trust region paradigm consists of iteratively
approximating the error function by a quadratic form in a neighbourhood
(region) centered around the current solution. The size of the region is
determined dynamically and depends on the quality of the current approximation.
We provide a complete description of the algorithm that we implemented together with a
comparison with the classical Newton method and with the Levenberg-Marquardt optimization algorithm.
\item Section 5 -- in order to check the validity of the solution thus obtained we compare
our results with the results obtained using the camera calibration method
proposed by Faugeras \& Toscani \cite{FaugerasToscani86}. This latter method estimates both the
intrinsic (interior) and extrinsic (exterior) parameters. We apply both methods
to the same data sets. We use the intrinsic
parameters determined by Faugeras \& Toscani's method as our camera model and
we compare the pose parameters determined by our method with the pose
parameters determined by the Faugeras \& Toscani method. We analyse the accuracy and
robustness of both our method and Faugeras-Toscani's method with respect to the number of correspondences, image noise, and
matching errors.
\end{itemize}

\section{Object pose from line correspondences}

We consider a pin-hole camera model and we assume that the parameters of the projection (the
intrinsic camera parameters) are known. The origin of the camera frame is at F -- the center of
projection, the $z$-axis is parallel to the optical axis, and the $xy$-plane is parallel with the
image plane. We assume that the optical axis is perpendicular onto the image plane.
An image point $m$ which has pixel coordinates $u$ and $v$ has the following
coordinates in the camera frame just described:
\begin{eqnarray*}
x & = & (u-u_{0})/\alpha _{u} \\
y & = & (v-v_{0})/\alpha _{v} \\
z & = & 1
\end{eqnarray*}
where $u_{0}$, $v_{0}$, $\alpha _{u}$, and $\alpha _{v}$ 
are the intrinsic parameters of the camera model \cite{Ganapathy84bis}
\cite{FaugerasToscani86} \cite{Tsai87}. Let $X$, $Y$, and $Z$ be the camera
coordinates of a space point $M$ which projects onto the image at $m$. We have:
\begin{equation}
\left\{	\begin{array}{c} x = \frac{X}{Z} \\ 
					 \\
			 y = \frac{Y}{Z} 
        \end{array} \right.
\label{eq:projection}
\end{equation}

\begin{figure*}
\centering
\includegraphics[width=10cm,height=8cm]{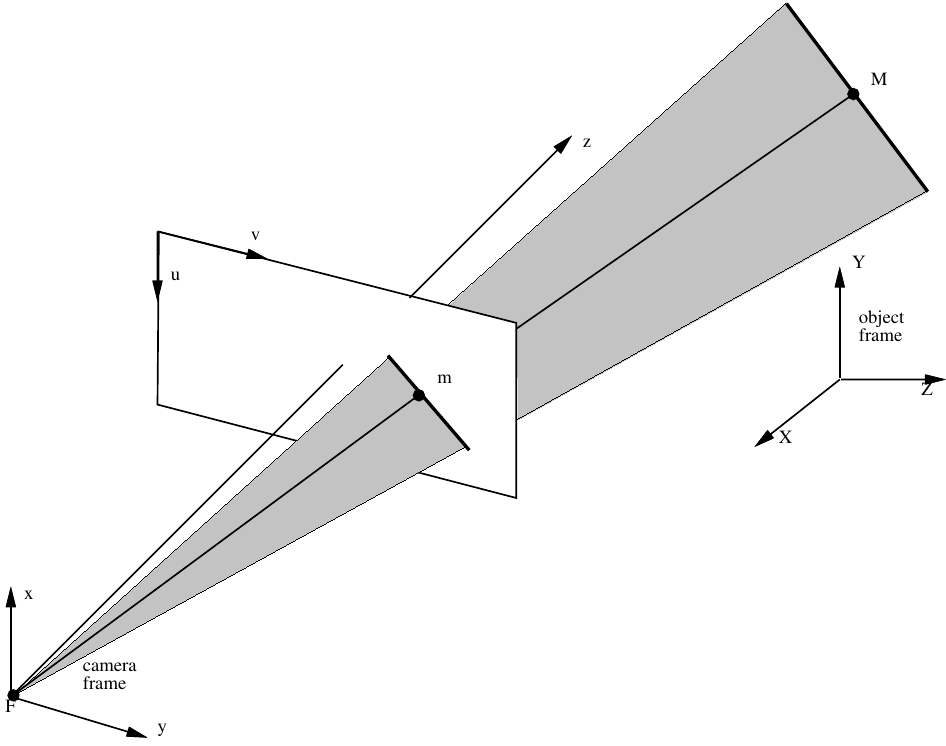}
\caption{The object line, its projection onto the image, and the center of projection $F$ are 
coplanar and this plane is shown in grey. $\vec{n}$ is the unit vector normal to this plane.}
\label{fig:geometry}
\end{figure*}

Moreover, the point $m$ is constrained to lie on an image line and the equation of this line is:
\begin{equation}
ax + by + c = 0
\label{eq:image-line}
\end{equation}
By substituting eq.~(\ref{eq:projection}) in eq.~(\ref{eq:image-line}) we obtain:
\begin{equation}
aX + bY + cZ = \vec{n} \cdot \vecteur{F}{M} = 0
\label{eq:interpretation-plane}
\end{equation}
which is the equation of the plane containing the center of projection, the image line and a
space point. $\vec{n}$ is the vector normal to this plane, e.g., Figure~\ref{fig:geometry}.

We consider now an object line. In the object frame this line is described parametrically by its
direction $\vec{v}$ and by a point vector $\vec{p}$ 
and it can be expressed in the camera frame as well:
\begin{eqnarray*}
\vec{p}\; ' & = & R \vec{p} + \vec{t} \\
\vec{v}\; ' &=& R\vec{v}
\end{eqnarray*}
where $\vec{p}\; '$ and $\vec{v}\; '$ describe the 3-D line in the camera
frame. The $3\times 3$ rotation matrix $R$ and the translation vector $\vec{t}$ describe the
rigid transformation from the object frame to the camera frame and are precisely the parameters
associated with the object pose problem. The correspondence constraints express the fact that
an object line belongs to the plane defined above, i.e., eq.~(\ref{eq:interpretation-plane}):
\begin{equation*}
\left\{  \begin{array}{c} \vec{n} \cdot \vec{v}\; ' = 0 \\ \vec{n} \cdot \vec{p}\; ' = 0 \end{array} \right.
\end{equation*}
or
\begin{equation}
\left\{  \begin{array}{c} \vec{n} \cdot (R\vec{v}) = 0 \\
		         \vec{n} \cdot (R \vec{p} + \vec{t}) = 0
	\end{array}
\right.
\label{eq:planarity-constraint}
\end{equation}

\begin{figure}
\centering
\includegraphics[width=0.49\textwidth]{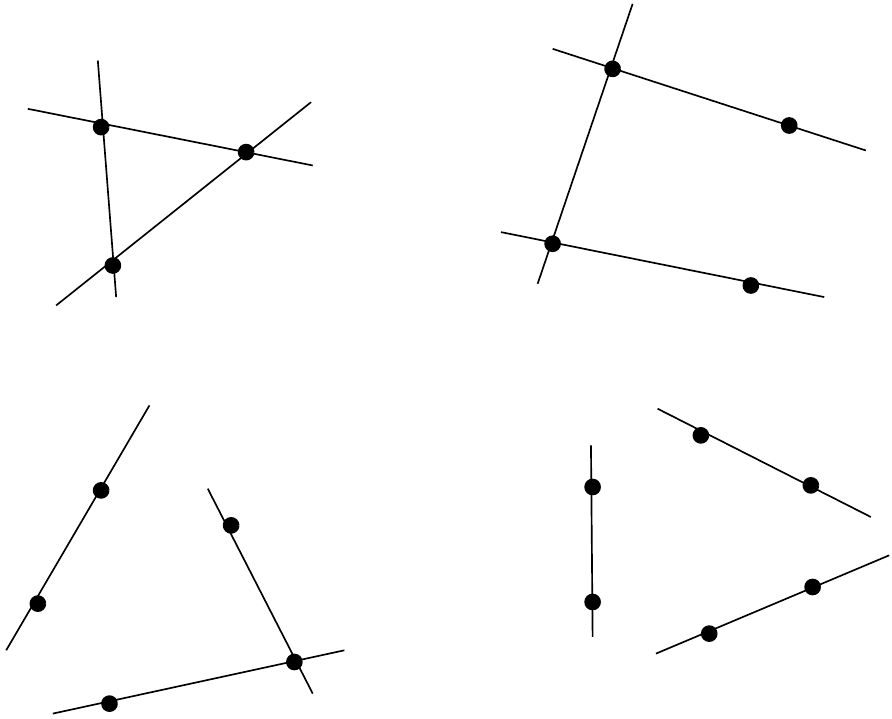}
\caption{Configurations of 3, 4, 5, or 6 points in general
position that can be replaced by 3 lines.}
\label{fig:lines-from-points}
\end{figure}

Therefore, each line correspondence provides 2 constraints. Since the number of unknown
parameters is 6 (3 for rotation and 3 for translation) the pose problem can be solved if at
least 3 line correspondences are available. In fact 3 lines can be built from 3, 4, 5, or 6
points, e.g., Figure~\ref{fig:lines-from-points}. More generally, $N$ line
correspondences can be built from 2$N$ point correspondences for $N\geq$3. 
Hence, the solution presented in this section is
a general solution for point and line correspondences for any number of correspondences. The minimal data set is either 3 points or 3 lines. In the
general case, if $N$ line correspondences are available, the pose problem becomes the problem of
solving for a set of $2N$ non linear constraints, or equivalently, the problem of minimizing the
following error function:
\begin{equation}
f(R,t) = \sum_{i=1}^{N}(\vec{n}_{i} \cdot (R\vec{v}_{i}))^{2} + \sum_{i=1}^{N}
         (\vec{n}_{i} \cdot (R\vec{p}_{i} + \vec{t}))^{2}
\label{eq:error-function-first}
\end{equation}

\subsection{Rotation, translation, and dual number quaternions}

It is well known since the French mathematician Chasles \cite{Chasles} that a
convenient parameterization of the displacement group (i.e., rigid
transformations) is a rotation about a unique axis not passing through the
origin and a translation along the same axis. Such a combination of rotation
and translation is called a {\em screw}. More formally, a screw is defined by
the triplet ($d$, $\phi$, $\cal L$), where:
\begin{itemize}
\item $d$ is the length of the translation along the screwing axis;
\item $\phi$ is the angle of rotation around this axis and
\item $\cal L$ is the screwing axis which may well be represented
parametrically by a direction unit vector $\vec{r}$ and a position vector
$\vec{l}$, with $\vec{r}\cdot\vec{l}=0$. 
\end{itemize}

A {\em dual number quaternion} is a mathematical representation of screws just
as a unit quaternion is a mathematical representation of rotations. The
interest of parameterizing screws (and hence rigid transformations) with dual
number quaternions resides in simple algebraic manipulations that are
available with quaternions and in simple and elegant expressions for pose
estimation problems. 

A dual number quaternion has a real part and a dual part:
\[
\hat{{\bf q}} = {\bf r} + \epsilon {\bf s}
\]
where {\bf r} and {\bf s} are quaternions and $\epsilon ^{2}=0$, \cite{WalkerShaoVolz92}.
\footnote{$\epsilon$ is known as the dual unity and obeys the following
multiplication rules:
0$\epsilon$=$\epsilon$0=0,1$\epsilon$=$\epsilon$1=$\epsilon$, and
$\epsilon ^{2}$=0.}

Now, a quaternion has a real part and three imaginary parts and may also be viewed as a 4-vector:
\[
{\bf r} = r_{0} + i r_{x} + j r_{y} + k r_{z} = (r_{0}, \vec{r}) = (r_{0} \; r_{x} \; r_{y} \; r_{z})^{T}
\]
Whenever $\hat{{\bf q}}$ represents a rigid transformation, its real and dual parts are defined
by:
\begin{eqnarray*}
{\bf r} & = & (\cos \frac{\phi}{2} \; , \; \; \sin \frac{\phi}{2} \vec{r}) \\
{\bf s} & = & (-\frac{d}{2} \sin \frac{\phi}{2}\; ,\; \; \frac{d}{2} \cos \frac{\phi}{2} \vec{r} +
	      \sin \frac{\phi}{2} \vec{l} \times \vec{r})
\end{eqnarray*}
One may easily verify that the following two constraints hold:
\begin{eqnarray*}
{\bf r} \cdot {\bf r} & = & 1 \\
{\bf r} \cdot {\bf s} & = & 0
\end{eqnarray*}
The rotation matrix $R$ and the translation vector $\vec{t}$ can be easily derived from {\bf r} and
{\bf s} using the following formulae:
\begin{equation}
\left( \begin{array}{cc} 1 & 0^{T} \\ 0 & R \end{array} \right) = W({\bf r})^{T} Q({\bf r})
\label{eq:rotation=quaternion}
\end{equation}
and
\begin{equation}
\left( \begin{array}{c} 0 \\ \vec{t} \end{array} \right) = 2 W({\bf r})^{T}{\bf s}
\label{eq:translation=quaternion}
\end{equation}
where $W({\bf r})$ and $Q({\bf r})$ are two 4$\times$4 matrices associated with a quaternion:
\begin{align}
\label{eq:matrix-Q}
Q({\bf r}) & = \left( \begin{array}{rrrr}
r_{0} & -r_{x} & -r_{y} & -r_{z} \\
r_{x} &  r_{0} & -r_{z} &  r_{y} \\
r_{y} &  r_{z} &  r_{0} & -r_{x} \\
r_{z} & -r_{y} &  r_{x} &  r_{0}
           \end{array}
    \right)
\\
\label{eq:matrix-W}
W({\bf r}) &= \left( \begin{array}{rrrr}
r_{0} & -r_{x} & -r_{y} & -r_{z} \\
r_{x} &  r_{0} &  r_{z} & -r_{y} \\
r_{y} & -r_{z} &  r_{0} &  r_{x} \\
r_{z} &  r_{y} & -r_{x} &  r_{0}
           \end{array}
    \right)
\end{align}

\subsection{The error function}

If 3-vectors are treated as purely imaginary quaternions, that is:
\[
{\bf v} = (0 \; , \; \vec{v})
\]
then the first constraint of eq.~(\ref{eq:planarity-constraint}) can be written as:
\begin{equation}
\vec{n} \cdot (R\vec{v})  =  {\bf n}^{T} (W({\bf r})^{T} Q({\bf r}){\bf v}) 
			 = {\bf r}^{T} Q({\bf n})^{T} W({\bf v}) {\bf r}
\label{eq:quadratic-rotation}
\end{equation}

Indeed, it is easy to verify that for two quaternions {\bf a} and {\bf b} we have the following
property:
\begin{equation}
Q({\bf a})  {\bf b} = W( {\bf b}) {\bf a}
\label{eq:Qab=Wba}
\end{equation}

The second constraint of eq.~(\ref{eq:planarity-constraint}) becomes:
\begin{eqnarray}
\vec{n} \cdot (R\vec{p} + \vec{t}) & = & {\bf n}^{T}(W({\bf r})^{T} Q({\bf r}){\bf p} + 2 W({\bf
r})^{T} {\bf s}) \nonumber \\
	& =& {\bf r}^{T} Q({\bf n})^{T} W({\bf p}) {\bf r} + {\bf r}^{T}(2Q({\bf n})^{T}){\bf s}
\label{eq:quadratic-rot-and-trans}
\end{eqnarray}

In other terms, each correspondence $i$ provides two quadratic constraints:
\[
{\bf r}^{T} A_{i} {\bf r} = 0
\]
and
\[
{\bf r}^{T} B_{i} {\bf r} + {\bf r}^{T} C_{i} {\bf s} = 0
\]
with ${\bf r}^{T}{\bf r}=1$, ${\bf r}^{T}{\bf s} = 0$, and $ A_{i}$, $B_{i}$, and
$C_{i}$ being three 4$\times$4 matrices:
\begin{eqnarray*}
A_{i} & = & Q({\bf n}_{i})^{T} W({\bf v}_{i}) \\
B_{i} & = & Q({\bf n}_{i})^{T} W({\bf p}_{i}) \\
C_{i} & = & 2Q({\bf n}_{i})^{T}
\end{eqnarray*}

We can now write a new expression of the error function associated with our problem, i.e.,
eq.~(\ref{eq:error-function-first}):
\begin{multline}
\label{eq:error-function-second}
f({\bf r},\; {\bf s}) = \sum_{i=1}^{N} \left(({\bf r}^{T} A_{i} {\bf r})^{2} + ({\bf r}^{T} B_{i} {\bf r} + {\bf r}^{T} C_{i} {\bf s})^{2}\right)  \\
+ \lambda ({\bf r}^{T}{\bf r}-1)^{2}
+ \lambda ({\bf r}^{T}{\bf s})^{2}
\end{multline}
where the parameters to be estimated are the elements of the quaternions (or 4-vectors)
{\bf r} and {\bf s}, and $\lambda$ is a positive number. $\lambda$ must be taken very large in order
to guarantee that the penalization constraints (${\bf r}^{T}{\bf r}=1$ and ${\bf r}^{T}{\bf s}=0$)
are satisfied. In practice we took $\lambda =50$ and we obtained very good results with this value.
However a too large value for $\lambda$ can make convergence quite inefficient (see Section~4).

Notice that an alternative to this error
function may be to consider the estimation of {\bf r} and {\bf s} separately. One may estimate the
rotation first using the following error function:
\begin{equation}
f(\bf r) = \sum_{i=1}^{N} ({\bf r}^{T} A_{i} {\bf r})^{2} + \lambda ({\bf r}^{T}{\bf r}-1)^{2}
\label{eq:error-function-rotation}
\end{equation}
Once the optimal value of {\bf r} is found, the computation of the optimal value of {\bf s} is
trivial since it is reduced to a linear least-square minimization problem:
\begin{equation}
f({\bf s}) = \sum_{i=1}^{N} ({\bf r}^{T} B_{i} {\bf r} + {\bf r}^{T} C_{i} {\bf s})^{2} + \lambda ({\bf r}^{T}{\bf s})^{2}
\label{eq:error-function-translation}
\end{equation}

\section{Object pose from point correspondences}

The error function that we devised in the previous section holds for line correspondences and for point
correspondences since a set of two quadratic constraints can be associated  either with one line or with
two points. In this section we devise an error function directly from point correspondences. This function has the
same structure -- sum of squares of quadratic constraints -- as with line correspondences,
provided that the rigid transformation is represented by a dual number quaternion. Each point
correspondence provides two constraints that express the collinearity between the object point, its image
projection, and the center of projection. 

Let again $M_{i}$ be an object point with coordinates $X_{i}$, $Y_{i}$, and $Z_{i}$ in an object frame and let $x_{i}$ and $y_{i}$
be the coordinates in the camera frame of its projection $m_{i}$. The collinearity between $M_{i}$, $m_{i}$, and $F$ (the center of projection) is
expressed by two equations \cite{Ganapathy84bis}, \cite{Haralicketal89}, e.g., Figure~\ref{fig:geometry}:
\begin{eqnarray}
\label{eq:x-collinearity}
x_{i} = \frac{r_{11} X_{i} + r_{12} Y_{i} + r_{13} Z_{i} + t_{1}}{r_{31} X_{i} + r_{32} Y_{i} + r_{33} Z_{i} + t_{3}} \\
\label{eq:y-collinearity}
y_{i} = \frac{r_{21} X_{i} + r_{22} Y_{i} + r_{23} Z_{i} + t_{2}}{r_{31} X_{i} + r_{32} Y_{i} + r_{33} Z_{i} + t_{3}}
\end{eqnarray}
where $r_{ij}$ is an element of the rotation matrix $R$ and $t_{i}$ is a component of the translation
vector $\vec{t}$. One may notice that the two constraints above can be written
more compactly as:
\begin{eqnarray}
\label{eq:x-collinearity-bis}
\left( \begin{array}{ccc} 1 & 0 & -x_{i} \end{array} \right)
R
\left( \begin{array}{c} X_{i} \\ Y_{i} \\ Z_{i} \end{array} \right)
+ 
\left( \begin{array}{ccc} 1 & 0 & -x_{i} \end{array} \right)
\left( \begin{array}{c} t_{1} \\ t_{2} \\ t_{3} \end{array} \right)
& = & 0 \\
\label{eq:y-collinearity-bis}
\left( \begin{array}{ccc} 0 & 1 & -y_{i} \end{array} \right)
R
\left( \begin{array}{c} X_{i} \\ Y_{i} \\ Z_{i} \end{array} \right)
+ 
\left( \begin{array}{ccc} 0 & 1 & -y_{i} \end{array} \right)
\left( \begin{array}{c} t_{1} \\ t_{2} \\ t_{3} \end{array} \right)
& = & 0
\end{eqnarray}

Let ${\bf M}_{i}$ be the purely imaginary quaternion associated with the point
vector $M_{i}$ and let ${\bf m}_{x_{i}}$ and ${\bf m}_{y_{i}}$ be two purely
imaginary quaternions each one associated with one ``half" of the point vector $m_{i}$:
\begin{align*}
{\bf M}_{i} & =  \left( 0 \; X_{i} \; Y_{i} \; Z_{i} \right)^{T} \\
{\bf m}_{x_{i}} & =  \left( 0 \; 1 \; 0 \; -x_{i} \right)^{T} \\
{\bf m}_{y_{i}} & =  \left( 0 \; 0 \; 1 \; -y_{i} \right)^{T}
\end{align*}
By substituting the rotation matrix $R$ and the translation vector
$\vec{t}$ with their dual number quaternion representation (eqs.~(\ref{eq:rotation=quaternion})
and (\ref{eq:translation=quaternion})) we obtain, for the
first constraint (see also eq.~(\ref{eq:Qab=Wba})):
\begin{align*}
{\bf m}_{x_{i}}^{T} & W({\bf r})^{T}Q({\bf r}){\bf M}_{i} + 2{\bf
m}_{x_{i}}^{T} W({\bf r})^{T}{\bf s} \\
& = ( W({\bf r}) {\bf m}_{x_{i}} )^{T}Q({\bf r}){\bf M}_{i} +
	2 (W({\bf r}) {\bf m}_{x_{i}})^{T} {\bf s} \\
&	= {\bf r}^{T} Q({\bf m}_{x_{i}})^{T} W ({\bf M}_{i}) {\bf r} +
	 2 {\bf r}^{T} Q({\bf m}_{x_{i}})^{T}{\bf s}
\end{align*}
There is a similar expression for the second constraint:
\begin{align*}
{\bf m}_{y_{i}}^{T} & W({\bf r})^{T}Q({\bf r}){\bf M}_{i} + 2{\bf
m}_{y_{i}}^{T}W({\bf r})^{T}{\bf s} \\
&=
{\bf r}^{T} Q({\bf m}_{y_{i}})^{T} W ({\bf M}_{i}) {\bf r} +
         2 {\bf r}^{T} Q({\bf m}_{y_{i}})^{T}{\bf s}
\end{align*}
Finally, equation~(\ref{eq:x-collinearity}) and equation~(\ref{eq:y-collinearity}), the collinear\-ity constraints, become:
\begin{eqnarray}
{\bf r}^{T} A_{i}^{p} {\bf r} + 2{\bf r}^{T} B_{i}^{p} {\bf s} = 0 \\
{\bf r}^{T} C_{i}^{p} {\bf r} + 2{\bf r}^{T} D_{i}^{p} {\bf s} = 0 
\end{eqnarray}
which can be written more compactly as two simple quadratic constraints:
\begin{eqnarray*}
\left(
\begin{array}{c}
{\bf r} \\ {\bf s}
\end{array}
\right)^{T}
\left(
\begin{array}{cc}
A_{i}^{p} & B_{i}^{p} \\
B_{i}^{p^{T}} & 0
\end{array}
\right)
\left(
\begin{array}{c}
{\bf r} \\ {\bf s}
\end{array}
\right) = 0 \\
\left(
\begin{array}{c}
{\bf r} \\ {\bf s}
\end{array}
\right)^{T}
\left(
\begin{array}{cc}
C_{i}^{p} & D_{i}^{p} \\
D_{i}^{p^{T}} & 0
\end{array}
\right)
\left(
\begin{array}{c}
{\bf r} \\ {\bf s}
\end{array}
\right) = 0
\end{eqnarray*}
where $A_{i}^{p}$, $B_{i}^{p}$, $C_{i}^{p}$, and $D_{i}^{p}$ are $4\times 4$ matrices given by:
\begin{eqnarray*}
A_{i}^{p} & = & Q({\bf m}_{x_{i}})^{T} W ({\bf M}_{i}) \\
B_{i}^{p} & = & Q({\bf m}_{x_{i}})^{T} \\
C_{i}^{p} & = & Q({\bf m}_{y_{i}})^{T} W ({\bf M}_{i}) \\
D_{i}^{p} & = & Q({\bf m}_{y_{i}})^{T}
\end{eqnarray*}

Hence, each point correspondence $i$ provides two quadratic constraints. To conclude, in the case of point correspondences,
the error function to be minimized is:
\begin{multline}
f({\bf r},\; {\bf s}) = \sum_{i=1}^{N} \left(
({\bf r}^{T} A_{i}^{p} {\bf r} + 2{\bf r}^{T} B_{i}^{p} {\bf s})^{2} +
({\bf r}^{T} C_{i}^{p} {\bf r} \right. \\
\left. + 2{\bf r}^{T} D_{i}^{p} {\bf s})^{2}\right) + \lambda ({\bf r}^{T}{\bf r}-1)^{2}
+ \lambda ({\bf r}^{T}{\bf s})^{2}
\label{eq:error-function-points}
\end{multline}

\section{The trust-region optimization method}

It is clear that the minimization of the error functions described by 
equations~(\ref{eq:error-function-second}), (\ref{eq:error-function-rotation}), and
(\ref{eq:error-function-points}) is
equivalent to the following non linear least squares problem:
\begin{equation}
   0 = \min _{x} \{ f(x)= \frac{1}{2} \sum_{j=1}^{m} \Phi_{j}^{2}(x): x \in I\!\!R^{n}\}
\label{eq:error-function}
\end{equation}
 where $\Phi_j(x)$ is twice continuously differentiable function from $I\!\!R^{n}$ to $I\!\!R$.
In our case $x$ is either a 4-vector (quaternion {\bf r} of eq.~(\ref{eq:error-function-rotation})) or
an 8-vector (the concatenation of quaternions {\bf r} and {\bf s} of
eqs.~(\ref{eq:error-function-second}) and (\ref{eq:error-function-points})).

There are  some methods which are usually used for dealing with unconstrained minimization problems, 
among which the steepest descent method, the Newton method, and Newton-like methods. 
The steepest descent method is both inefficient and unreliable because of its slow rate of linear 
convergence. The Newton method converges rapidly but its main drawback is that the method  is only well 
defined under certain conditions and even in these cases it does not ensure the descent. 
In this section we proceed to present a Newton-like method which is generally applicable and 
globally convergent while it retains the rapid convergence rate of Newton's method -- the trust region method.

\begin{figure*}[t!]
\centering
\includegraphics[width=10cm,height=5cm]{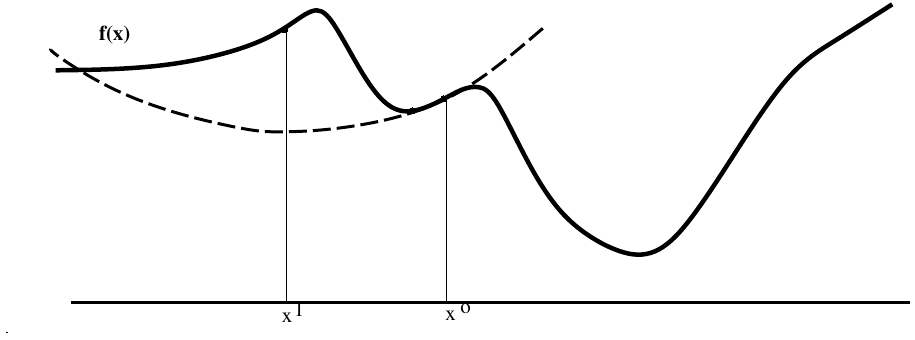}
\caption{An example of minimization when several minima are present. Newton's method may lead to an increase of the value of the function $f(x)$.}
\label{fig:newton}
\end{figure*}

\begin{figure*}[t!h!]
\centering
\includegraphics[width=10cm,height=5cm]{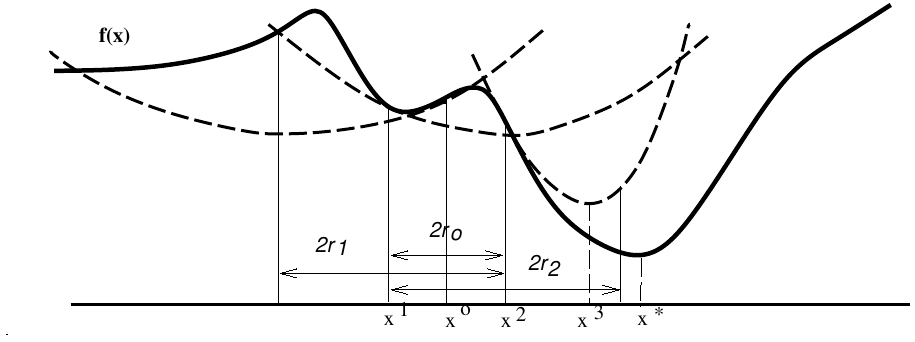}
\caption{This figure shows an example of minimizing the function $f(x)$ with the trust region method
which always ensures that at each iteration the value of the function decreases.}
\label{fig:trustregion}
\end{figure*}

In order to understand the difference between Newton's method (which has been so far widely used in
computer vision) and trust region methods we consider a simple example.
Figure~\ref{fig:newton} shows a single-valued function $f(x)$ to be minimized and which has more than
one minimum. From an initial point $x^{0}$, Newton's method approximates the function with a parabola by
using a second order Taylor's expansion of $f(x)$ around $x^{0}$. The next point, $x^{1}$, corresponds
to the minimum of that parabola but this value increases $f(x)$. The idea of the trust-region method is
to approximate the function around the current iterate (say $x^{0}$) with a parabola as well and to
search for the minimum of that parabola in a {\em restricted} set around $x^{0}$ rather than over the
entire definition set of $f(x)$. Figure~\ref{fig:trustregion} shows an example of applying this method
to the same function as above. At $x^{0}$ the function is approximated with a parabola on an interval of
size $2r_{0}$. The minimum of the parabola over this interval is at $x^{1}=x^{0}-r_{0}$. At $x^{1}$ the
function is approximated with a parabola again, but on an interval of increased size $2r_{1}$, with
$r_{1}=2r_{0}$. The
minimum of the new parabola is at $x^{2}=x^{1}+r_{1}$. From now on, the behaviour of the algorithm is
very similar to the Newton's method and it will ultimately converge to the global minimum $x^{*}$.


Denote by $\nabla f(x)$ and  $\nabla ^{2}f(x)$ the  gradient  and the Hessian of a function $f(x)$.  From eq. (\ref{eq:error-function}) it is easy to verify that:
\begin{eqnarray}
\label{eq:the-gradient}
 \nabla f(x)&=& \sum_{j=1}^{m} \Phi_{j}(x)\nabla\Phi_{j}(x) =J(x)^{T}\Phi(x), \\
 \nabla^{2}f(x) &=& \sum_{j=1}^{m}\nabla\Phi_{j}(x)\nabla\Phi_{j}(x)^{T}+ \sum_{j=1}^
{m} \Phi_{j}(x) \nabla^{2}\Phi_{j}(x) \nonumber \\
 \label{eq:the-hessian}
               &=& J(x)^{T}J(x) + \sum_{j=1}^{m} \Phi_{j}(x)\nabla^{2}\Phi_{j}(x)
\end{eqnarray}
where $\Phi(x)=(\Phi _{1}(x), \ldots, \Phi _{m}(x))^{T}$ and 
\[    J(x)=(\nabla\Phi _{1}(x),\ldots,\nabla\Phi _{m}(x))^{T}\]
is the $m \times n $ Jacobian matrix of $\Phi(x)$.  The Hessian of such a least-square error function is
a combination of first- and second-order terms:
\[
\nabla^{2}f(x) = J(x)^{T}J(x) + Q(x)
\]
with: $Q(x)=\sum_{j=1}^{m} \Phi_{j}(x)\nabla^{2}\Phi_{j}(x)$. In practice, Gauss-Newton
approximation of the Hessian is used, $H(x) = J(x)^{T}J(x)$. This is based on the premise
that the first-order term will eventually dominate the second-order term, $Q(x)$
\cite{GillMurrayWright89}.

Let $x^{k}$ denote the current estimate of the solution; a vector over scripted by $k$ will
denote a vector evaluated at the $k^{th}$ iteration of the algorithm. 
A scalar subscripted by $k$ will denote a quantity evaluated at the $k^{th}$ iteration.
The basic idea of the
trust-region optimization method consists of successively approximating the error function by a
{\em local quadratic form} (or a paraboloid) in a neighbourhood of the current solution $x^{k}$:
\[ f(x^{k}+d) \approx f(x^{k})+ q_{k}(d), \; \; \|d\|\leq \delta_{k} \]
where:
\begin{equation}
q_{k}(d) = \nabla f(x^{k})^{T}d + \frac{1}{2} d^{T} H(x^{k})d
\label{eq:the-local-quadratic-form}
\end{equation}

The error function will be reduced via the direction $d^{k}$, i.e., $x^{k+1}=x^{k}+d^{k}$, where
$d^{k}$ is a  minimizer of  the local quadratic form over a restricted spherical
region centered around $x^{k}$: {\em the trust region}:
\begin{equation}
d^{k} \in \mbox{arg} \min _{d} \{ q_{k}(d) : \|d\| \leq \delta _{k} \}
\label{eq:lqf-minimization}
\end{equation}
The parameter $\delta _{k}$ is called the trust radius and is determined dynamically using a
measure of the quality of the approximation; this is measured by a quality coefficient $r_{k}$:
\begin{equation}
   \label{eq:quality-coefficient}
    r_{k}= \frac{f(x^{k})-f(x^{k}+d^{k})}{ q_{k}(0)-q_{k}(d^{k})}
\end{equation}
which measures the ratio between the
{\em actual reduction} of the error function when moving from $x^{k}$ to
$x^{k}+d^{k}$ and the {\em predicted reduction} according to the local quadratic approximation.
If $r_{k}$ is too small it means that the approximation is not good and the trust region should
be decreased. Otherwise the trust region should be increased. Roughly, the following strategy
will be applied:
\begin{itemize}
\item if $r_{k} \geq r$ (with $0<r\leq 1$) then the new iterate is accepted, i.e.,
$x^{k+1}=x^{k}+d^{k}$ and in addition the trust region is increased;
\item if $r_{k} < r$ then the iterate is unchanged and in addition the trust region is
decreased. This process is repeated until the local quadratic form yields a satisfactory
approximation inside the trust region (which will necessarily occur for small radii since the
gradient is supposed to be exactly known and the first-order term in $q_{k}$ 
becomes dominant if $\|g_{k} \| \neq 0$).
\end{itemize}

The local quadratic form depends on the gradient and the Hessian of the error function. Hence,
the minimum thus found has ``good" second-order properties. In a trust-region method the main
difficulty resides in the minimization of the local quadratic form. Various trust region
algorithms differ upon the method being used to minimize the local quadratic form inside the
trust region. A detailed discussion on the theoretical aspects of this problem is given in Appendix~A.

We propose to apply the following practical trust-region algorithm to our problem (see also
Clermont \& al. \cite{kn:Cler} and Tao \& al. \cite{kn:Tao}).
The idea is to use the Gauss-Newton's approximation of the Hessian for determining a 
search direction. Generally, $H(x)=J^T(x)J(x)$ is positive semi-definite and we can make it positive definite by considering
\begin{equation*}           H'(x)=H(x)+ \alpha I \end{equation*}
where $\alpha$ is a chosen small positive number (for example $\alpha=0.001$).  So $H'(x)+\mu I \geq 0$ (condition (i) in Theorem 1 (Appendix~A)) holds for all $\mu \geq 0$ .  To solve  (\ref{eq:lqf-minimization}) we first check whether 
$\mu_*=0$ when solving $H'(x^k)d=-\nabla f(x^k)$.  If the obtained solution $d^k$ satisfies $\|d^k\|\leq \delta_k$ then $d^k$ is an optimal solution. Otherwise, $\mu_* >0$ and an unique solution to (\ref{eq:lqf-minimization}) can be found by  a very efficient algorithm proposed by Hebden (see Appendix~A).  

\subsection{The practical trust-region algorithm}

The trust region algorithm is summarized as follows:

\begin{itemize}
 \item {\em Initialization} :  Choose an initial iterate $x^o$, a trust radius $ \delta_{o}$ and small positive numbers $\epsilon,\epsilon_g,\epsilon_f$.
 Set k=0.
 
 \item {\em Iteration k=0,1,\ldots }
  \begin{description}
    \item {k.1} Compute $f_{k}=f(x^{k}),g_{k}=\nabla f(x^{k})$ and $H_{k}=J^T(x^k)J(x^k)$.
    \item {k.2} If $\|g_{k}\| \leq \epsilon_{g}$ or $\delta_{k} \leq \epsilon_{\delta
}$ or $f_{k}\leq \epsilon_{f}$ then stop: $x^{k}$ is a solution.
    \item {k.3} Consider $H'_k=H_k+0.001 I$. Let $d$ be a solution of the system
\[
      H'_{k}d=- g_{k}
\]
   If $\|d\| < \delta_{k} - \epsilon$ then $d^{k}=d$. Otherwise, use Hebden's algorithm 
to obtain $d^k$.
    \item {k.4} Compute $r_{k}$ using eq. (\ref{eq:quality-coefficient}).
    \item {k.5} If $r_{k} \geq r$ then
         \[ x^{k+1} := x^{k}+d^{k} \]
          if $r_{k} \geq s$ then
        \[ \delta_{k+1}:=2\delta_{k} \]
          otherwise
           \[ \delta_{k+1}:=\delta_{k} \]
       Set $k:=k+1$ and return to k.1
     \item {k.6} If $r_{k} < r$ then
          \[ \delta_{k}:=\delta_{k}/2 \]
   and return to k.3.
  \end{description}
\end{itemize}

The parameter $r$ must belong to the interval
[0.1, 0.3] and the parameter $s$ must belong to the interval [0.5, 0.8]. For our
application, these parameters were set at: $r=0.25$ and $s=0.75$.

\subsection{Comparison with Levenberg-Marquardt}
A well known non-linear optimization method that could have been used for solving our problem is
a method that is due to Levenberg and Marquardt \cite{kn:Mo78}, \cite{Fletcher90}. 
This method uses the Gauss-Newton
approximation of the Hessian and the $k^{th}$ search direction is defined as the solution of:
\[ (J(x^{k})^{T}J(x^{k})+\nu_{k}I)d^{k}=-J(x^{k})^{T}\Phi(x^{k}) \]
where $\nu_{k}$ is a non-negative scalar. It can be shown that, for {\em some} scalar 
$\delta_{k}$ related to $\nu_{k}$, the vector $d^{k}$ is the solution of the 
constrained subproblem:
\[  \min _{d} \{\frac{1}{2} \|J(x^{k})d+\Phi(x^{k})\|^{2},\; : \|d\| \leq \delta_{k}. \} \]
which is similar to the minimization of the local quadratic form just described.  The Levenberg-Marquardt method does have disadvantage when $\nu_k$ is close to zero in which case $d^k$ is badly determined by $\nu_k$. 

Both our method and Levenberg-Marquardt's one are based on the assumption that ultimately
$J(x^{k})^{T}J(x^{k})$ is a good approximation of the Hessian, i.e., that
the second-order term in eq.~(\ref{eq:the-hessian}) 
can be neglected. This assumption is not justified for the
so-called {\em large-residual} problems in which the optimal value $\|f(x^{*})\|$ 
is not ``small''. In this case, we must use the exact Hessian which is no longer positive and
semi-definite. Therefore one cannot use the Levenberg-Marquardt method any more. 
The trust-region method that we just described can be used by replacing the step k.3 (Hebden's
algorithm)
with the {\em safeguarding algorithm} proposed by Mor\'{e} \& Sorensen \cite{kn:MoSo79}. 
However, the latter algorithm is more time consuming than the Hebden's algorithm.

\section{Experimental results}

The trust region algorithm is particularly well-suited for solving the object pose from a single
view problem because the error function is a sum of squares of quadratic constraints. 
Indeed, the trust region algorithm -- generally applicable for any non linear constraints -- is
more robust and more efficient when these constraints are quadratic. This explains why the algorithm
can attain the global solution independently of the initialisation in a very small number of
iterations. This is also one reason for which dual number quaternions are a good representation for
rotation and translation.
The experiments that we performed can be summarized as follows:
\begin{itemize}
\item A 3-D calibrating object with 460 calibrating points is viewed by a camera and point-to-point
correspondences are established. The accuracy of the 3-D positions of these points is within 1/10 of
a millimeter and the accuracy of the image points is within 1/10 of a pixel;
\item The intrinsic and extrinsic camera parameters are determined using these 460 point
correspondences and the linear method of Faugeras \& Toscani \cite{FaugerasToscani86}. This solution
is taken to be the ``ground-truth". The intrinsic camera parameters thus found will be used 
throughout the experiments with the non-linear sol\-ver described in this paper. The numerical
values of these intrinsic parameters are displayed on the last row of
Table~\ref{table:faugeras-points}. The numerical values of the extrinsic parameters are as follows:
	\begin{itemize}
	\item The axis of rotation $\vec{r}$: (--0.185148 --0.973420 --0.156567)$^{T}$
	\item The angle of rotation $\phi$ (radians): 2.9512
	\item The translation vector $\vec{t}$ (in centimeters): (2.627650 --11.006138 77.479236)$^{T}$
	\end{itemize}
\item Subsets of point correspondences are randomly selected from the initial set of 460 points.
These subsets contain 10, 20, 40, 60, 80, 100, 200, 300, and 400 points. Hence, sets of 5, 10, 20, 30, 40,
50, 100, 150, and 200 line correspondences are also available;

\begin{small}
\begin{table*}[h]
\begin{center}
\begin{tabular}{||c|c|c|c|c|c|c|c|c|c|c||}
\hline
     Number 
     & \multicolumn{3}{c|}{Transformation error}
     & \multicolumn{3}{c|}{Image error} 
     & \multicolumn{4}{c||}{Intrinsic parameters}\\
\cline{2-11}
of points&axis&angle&translation&largest&smallest&average&$u_{0}$&$v_{0}$&$\alpha _{u}$&$\alpha _{v}$\\ 
\hline
10&0.0170&0.0032&2.4277&0.9754&0.0008&0.2150&243.1&224.3&-1677.8&1183.2 \\
20&0.0064&0.0143&3.2806&1.0983&0.0002&0.1912&269.0&233.2&-1663.5&1173.5 \\
40&0.0016&0.0047&0.8157&1.0935&0.0018&0.1577&235.2&238.3&-1715.7&1209.6 \\
60&0.0009&0.0015&0.2563&0.7255&0.0007&0.1508&239.9&242.9&-1734.2&1223.2 \\
80&0.0024&0.0038&0.6620&0.9277&0.0001&0.1500&235.9&245.5&-1741.8&1229.0 \\
100&0.0020&0.0003&0.1727&0.6431&0.0006&0.1500&245.6&238.8&-1730.3&1221.0 \\
200&0.0011&0.0001&0.2249&0.7806&0.0001&0.1444&243.6&243.5&-1734.6&1223.8 \\
300&0.0015&0.0016&0.3334&0.7471&0.0010&0.1425&246.0&244.1&-1736.8&1225.1 \\
400&0.0002&0.0001&0.0147&0.6979&0.0000&0.1444&244.1&241.2&-1730.1&1220.5 \\
460&0.0000&0.0000&0.0000&0.6987&0.0006&0.1446&243.8&241.2&-1730.3&1220.7 \\
\hline
\end{tabular}
\caption{This table shows the performance of the linear method of Faugeras and
Toscani as a function of the number of point correspondences. The solution
obtained with 460 points (the last row)
is taken to be the ground-truth. While rotation seems
to be relatively stable, translation degrades gently as the number of points
decreases. Notice however the large variations of the intrinsic parameters.}
\label{table:faugeras-points}
\end{center}
\end{table*}
\end{small}

\begin{small}
\begin{table*}[h]
\begin{center}
\begin{tabular}{||c|c|c|c|c|c|c|c|c|c|c||}
\hline
     Noise max. 
     & \multicolumn{3}{c|}{Transformation error}
     & \multicolumn{3}{c|}{Image error} 
     & \multicolumn{4}{c||}{Intrinsic parameters}\\
\cline{2-11}
amplitude&axis&angle&translation&largest&smallest&average&$u_{0}$&$v_{0}$&$\alpha _{u}$&$\alpha _{v}$\\ 
\hline
0.00&0.0023&0.0006&0.3333&0.5997&0.0008&0.1461&244.0&236.5&-1726.2&1217.6\\
0.01&0.0042&0.0035&0.6013&1.1291&0.0002&0.1574&235.2&246.8&-1738.3&1225.9\\
0.05&0.0014&0.0006&0.0970&0.7807&0.0011&0.1460&245.2&242.1&-1731.9&1221.9\\
0.10&0.0018&0.0029&0.4556&0.8290&0.0009&0.1463&239.4&238.7&-1721.9&1214.7\\
0.50&0.0103&0.0046&0.9617&0.7590&0.0016&0.1589&241.8&230.2&-1714.2&1208.9\\
1.00&0.0198&0.0177&11.4770&2.2091&0.0022&0.3827&222.0&198.4&-1478.2&1042.3\\
1.50&0.0414&0.0306&17.8853&3.3508&0.0028&0.6589&226.5&159.7&-1336.2&938.9\\
2.00&0.0460&0.0578&23.1873&4.3128&0.0022&1.0022&197.4&152.6&-1218.2&856.5\\
3.00&0.1774&0.0583&44.3465&13.9919&0.0368&2.6464&194.2&156.3&-734.1&510.7\\
4.00&1.2958&0.0339&55.6320&32.0500&0.0171&5.6593&160.2&141.1&-441.4&303.9\\
5.00&0.7169&0.0753&61.1560&45.8325&0.0312&7.9906&201.9&185.8&-336.3&231.2\\
\hline
\end{tabular}
\caption{This table shows the performance of the linear method (Faugeras and
Toscani) for 100 point correspondences and in the presence of image noise.
The first column of this table shows the maximum noise amplitude around the
nominal image point locations. Rotation resists quite nicely to noise
while translation is not reliable for a maximum noise amplitude larger than 
half of a pixel.}
\label{table:faugeras-noise}
\end{center}
\end{table*}
\end{small}

\begin{small}
\begin{table*}[h]
\begin{center}
\begin{tabular}{||c|c|c|c|c|c|c|c|c||}
\hline
     Number 
     & \multicolumn{3}{c|}{Transformation error}
     & \multicolumn{3}{c|}{Image error} 
     & \multicolumn{2}{c||}{Performance}\\
\cline{2-9}
of lines&axis&angle&translation&largest&smallest&average&iterations&CPU time\\ 
\hline
   5 &0.0034& 0.0005& 0.0056&0.9421&0.0106&0.2870&10&0.12\\
  10& 0.0047& 0.0008& 0.0156&0.9049&0.0098&0.2127& 7&0.03\\
  20& 0.0042& 0.0012& 0.0037&0.6071&0.0020&0.1804& 6&0.02\\
  30& 0.0038& 0.0004& 0.0038&0.7150&0.0117&0.1775& 6&0.03\\
  40& 0.0036& 0.0001& 0.0025&0.6534&0.0060&0.1856& 7&0.05\\
  50& 0.0036& 0.0005& 0.0042&0.7525&0.0105&0.1872& 6&0.07\\
 100 &0.0037& 0.0001& 0.0013&0.6784&0.0022&0.1702& 6&0.12\\
 150& 0.0036& 0.0003& 0.0001&0.6067&0.0078&0.1667& 6&0.17\\
 200& 0.0037& 0.0004& 0.0000&0.5468&0.0098&0.1658& 6&0.22\\
 230& 0.0035& 0.0004& 0.0004&0.6010&0.0086&0.1669& 6&0.25\\
\hline
\end{tabular}
\caption{The performance of the rotation-then-translation implementation of
our method with line correspondences. The CPU time is in seconds and was
obtained on a Sun/Sparc10.}
\label{table:lines-RthenT}
\end{center}
\end{table*}
\end{small}

\begin{small}
\begin{table*}[h]
\begin{center}
\begin{tabular}{||c|c|c|c|c|c|c|c|c||}
\hline
     Noise max. 
     & \multicolumn{3}{c|}{Transformation error}
     & \multicolumn{3}{c|}{Image error} 
     & \multicolumn{2}{c||}{Performance}\\
\cline{2-9}
amplitude&axis&angle&translation&largest&smallest&average&iterations&CPU time\\ 
\hline
  0.00& 0.0034& 0.0005& 0.0001&0.5653&0.0121&0.1675& 6&0.07\\
  0.01& 0.0039& 0.0005& 0.0007&0.6794&0.0105&0.1729 &6&0.07\\
  0.05& 0.0038& 0.0010& 0.0010&0.5823&0.0126&0.1696& 6&0.07\\
  0.10& 0.0039& 0.0003& 0.0016&0.7211&0.0025&0.1732& 6&0.07\\
  0.50& 0.0034& 0.0005& 0.0028&0.7690&0.0129&0.2016& 7&0.07\\
  1.00& 0.0034& 0.0004& 0.0024&0.6709&0.0038&0.2620& 6&0.05\\
  1.50& 0.0040& 0.0008& 0.1024&1.7369&0.0470&0.5088& 6&0.05\\
  2.00& 0.0038& 0.0001& 0.1326&1.6247&0.0276&0.5777& 6&0.07\\
  3.00& 0.0042& 0.0014& 0.4713&2.9518&0.0805&1.1266& 6&0.07\\
  4.00& 0.0033& 0.0013& 0.1499&2.0029&0.0162&0.7612& 6&0.07\\
  5.00& 0.0034& 0.0015& 0.1055&2.2339&0.3024&1.1693& 6&0.05\\
\hline
\end{tabular}
\caption{Same as the table before but the quantities are 
functions of image noise.}
\label{table:lines-RthenT-noise}
\end{center}
\end{table*}
\end{small}

\begin{small}
\begin{table*}[h]
\begin{center}
\begin{tabular}{||c|c|c|c|c|c|c|c|c||}
\hline
     Number 
     & \multicolumn{3}{c|}{Transformation error}
     & \multicolumn{3}{c|}{Image error} 
     & \multicolumn{2}{c||}{Performance}\\
\cline{2-9}
of lines&axis&angle&translation&largest&smallest&average&iterations&CPU time\\ 
\hline
  5&0.0151&0.0316&1.7323&6.3106&0.1415&2.9870&3&0.05\\
 10&0.0036&0.0018&0.0014&0.6187&0.0063&0.1948&18&0.13\\
 20&0.0040&0.0021&0.0023&0.5790&0.0055&0.1830&30&0.33\\
 30&0.0035&0.0021&0.0032&0.7203&0.0134&0.1933&23&0.35\\
 40&0.0034&0.0007&0.0018&0.6408&0.0132&0.1881&23&0.45\\
 50&0.0034&0.0004&0.0026&0.7202&0.0049&0.1890&24&0.53\\
 100&0.0037&0.0002&0.0017&0.6992&0.0052&0.1726&14&0.65\\
 150&0.0037&0.0002&0.0004&0.6320&0.0094&0.1686&19&1.28\\
 200&0.0035&0.0003&0.0004&0.5415&0.0054&0.1666&8&0.57\\
 230&0.0036&0.0003&0.0009&0.6380&0.0078&0.1680&8&0.67\\
\hline
\end{tabular}
\caption{Performance of the rotation-and-translation implemetation of the
trust region method for line correspondences.}
\label{table:lines-RandT}
\end{center}
\end{table*}
\end{small}

\begin{small}
\begin{table*}[h]
\begin{center}
\begin{tabular}{||c|c|c|c|c|c|c|c|c||}
\hline
     Noise max. 
     & \multicolumn{3}{c|}{Transformation error}
     & \multicolumn{3}{c|}{Image error} 
     & \multicolumn{2}{c||}{Performance}\\
\cline{2-9}
amplitude&axis&angle&translation&largest&smallest&average&iterations&CPU time\\ 
\hline
0.00& 0.0033& 0.0008& 0.0003&0.4916&0.0096&0.1706&11&0.27\\
0.01& 0.0038& 0.0000& 0.0004&0.6495&0.0045&0.1720&21&0.48\\
0.05& 0.0041& 0.0008& 0.0028&0.6654&0.0031&0.1768&22&0.55\\
0.10& 0.0037& 0.0007& 0.0009&0.6408&0.0086&0.1690&21&0.53\\
0.50& 0.0040& 0.0030& 0.0050&1.0173&0.0128&0.2381&10&0.20\\
1.00& 0.0057& 0.0028& 0.0421&1.3253&0.0055&0.3742&34&0.87\\
1.50& 0.0033& 0.0010& 0.0553&1.4390&0.0206&0.4949&22&0.52\\
2.00& 0.0039& 0.0081& 0.0798&1.6826&0.0354&0.6471&21&0.52\\
3.00& 0.0032& 0.0083& 0.3092&2.9927&0.0842&1.0871&25&0.67\\
4.00& 0.0048& 0.0213& 0.6658&3.3989&0.0473&1.1541&22&0.57\\
5.00& 0.0065& 0.0222& 0.5787&4.0622&0.0557&1.3912&22&0.52\\
\hline
\end{tabular}
\caption{Same as the table before but the quantities are
functions of image noise.}
\label{table:lines-RandT-noise}
\end{center}
\end{table*}
\end{small}

\begin{small}
\begin{table*}[h]
\begin{center}
\begin{tabular}{||c|c|c|c|c|c|c|c|c||}
\hline
     Number 
     & \multicolumn{3}{c|}{Transformation error}
     & \multicolumn{3}{c|}{Image error} 
     & \multicolumn{2}{c||}{Performance}\\
\cline{2-9}
of points&axis&angle&translation&largest&smallest&average&iterations&CPU time\\ 
\hline
  10&0.0032& 0.0010& 0.0171&0.6706&0.0015&0.2186&24&0.32\\
  20&0.0033& 0.0012& 0.0001&0.5503&0.0051&0.1780&10&0.18\\
  40&0.0035& 0.0000& 0.0000&0.5747&0.0189&0.1709& 9&0.37\\
  60&0.0035& 0.0002& 0.0002&0.5220&0.0115&0.1665& 9&0.45\\
  80&0.0035& 0.0000& 0.0002&0.5201&0.0163&0.1657& 9&0.55\\
 100&0.0035& 0.0003& 0.0001&0.5747&0.0112&0.1675& 9&0.70\\
 200&0.0036& 0.0003& 0.0000&0.5385&0.0109&0.1652& 9&1.62\\
 300&0.0035& 0.0003& 0.0004&0.5168&0.0054&0.1666&10&2.88\\
 400&0.0035& 0.0002& 0.0002&0.5385&0.0158&0.1657&10&4.00\\
 460&0.0035& 0.0001& 0.0001&0.5427&0.0228&0.1654&10&4.83\\
\hline
\end{tabular}
\caption{The  performance of the rotation-and-translation implementation of
the trust region method with point correspondences. This implementation seems
to provide the most accurate results and the most stable ones with respect to the
number of correspondences.}
\label{table:points-RandT}
\end{center}
\end{table*}
\end{small}

\begin{small}
\begin{table*}[h]
\begin{center}
\begin{tabular}{||c|c|c|c|c|c|c|c|c||}
\hline
     Noise max. 
     & \multicolumn{3}{c|}{Transformation error}
     & \multicolumn{3}{c|}{Image error} 
     & \multicolumn{2}{c||}{Performance}\\
\cline{2-9}
amplitude&axis&angle&translation&largest&smallest&average&iterations&CPU time\\ 
\hline
0.00&0.0034& 0.0004& 0.0012&0.4761&0.0121&0.1708& 9&0.77\\
0.01&0.0035& 0.0003& 0.0008&0.5356&0.0091&0.1678& 9&0.72\\
0.05&0.0035& 0.0000& 0.0001&0.5414&0.0096&0.1656& 9&0.80\\
0.10&0.0038& 0.0004& 0.0004&0.5877&0.0105&0.1664& 9&0.72\\
0.50&0.0034& 0.0002& 0.0001&0.5635&0.0069&0.1823& 9&0.80\\
1.00&0.0039& 0.0005& 0.0002&0.6765&0.0108&0.2026& 9&0.78\\
1.50&0.0036& 0.0023& 0.0024&0.7637&0.0062&0.2046& 9&0.80\\
2.00&0.0047& 0.0114& 0.0014&1.3642&0.0173&0.4545& 9&0.70\\
3.00&0.0059& 0.0020& 0.0482&1.0298&0.0289&0.3472& 9&0.80\\
4.00&0.0043& 0.0004& 0.0001&1.0785&0.0214&0.3643&10&0.93\\
5.00&0.0070& 0.0078& 0.3820&2.3401&0.0551&0.8375&10&0.83\\
\hline
\end{tabular}
\caption{Same as the table before but the quantities are functions of image noise.}
\label{table:points-RandT-noise}
\end{center}
\end{table*}
\end{small}

\item The linear method of Faugeras and Toscani is applied to these subsets of point correspondences.
The solution obtained with each subset is then compared to the ground-truth, e.g.,
Table~\ref{table:faugeras-points}. For one of these point subsets (100 points) noise with increasing
maximum amplitude is added to the image point locations. Again, the solutions thus obtained are
compared with the ground-truth solution, e.g., Table~\ref{table:faugeras-noise}.

\item The trust region algorithm is applied both to sets of line correspondences (Section~2) and
to point correspondences (Section~3). Moreover, there are two cases for line correspondences: (i)
determine the rotation first with the non linear algorithm and then determine the translation with a
linear algorithm and (ii) determine the rotation and the translation simultaneously with the non
linear algorithm. Hence, there are three implementations of the non linear algorithm:
\begin{enumerate}
\item Rotation {\em then} translation with {\em line} correspondences, Table~\ref{table:lines-RthenT}
and Table~\ref{table:lines-RthenT-noise};
\item Rotation {\em and} translation with {\em line} correspondences,
Table~\ref{table:lines-RandT} and Table~\ref{table:lines-RandT-noise}, and
\item Rotation {\em and} translation with {\em point} correspondences,
Table~\ref{table:points-RandT} and Table~\ref{table:points-RandT-noise}.
\end{enumerate}
Each one of the implementations above is applied to subsets of (points or lines) noise-free data and
to data degraded by noise (100 points or 50 lines). Degraded data is obtained
by adding uniformly distributed random noise (with a predefined maximum 
amplitude) to the image location of every point in a data set.
Each solution thus obtained is compared to 
the ground-truth solution. 
\end{itemize}

The discrepancy between a solution $s$ and the ground-truth solution $s_{0}$ is measured in
``transformation space". The transformation errors are given by:
\begin{eqnarray*}
e_{axis} & =&  \| \vec{r} - \vec{r}_{0} \| \\
e_{angle}& =& | \phi - \phi _{0} | \\
e_{translation}& =&  \| \vec{t} - \vec{t}_{0} \|
\end{eqnarray*}
where quantities subscripted by 0 correspond to the ground-truth solution. 

\begin{figure*}[t!h!]
\centering
\includegraphics[width=0.49\textwidth]{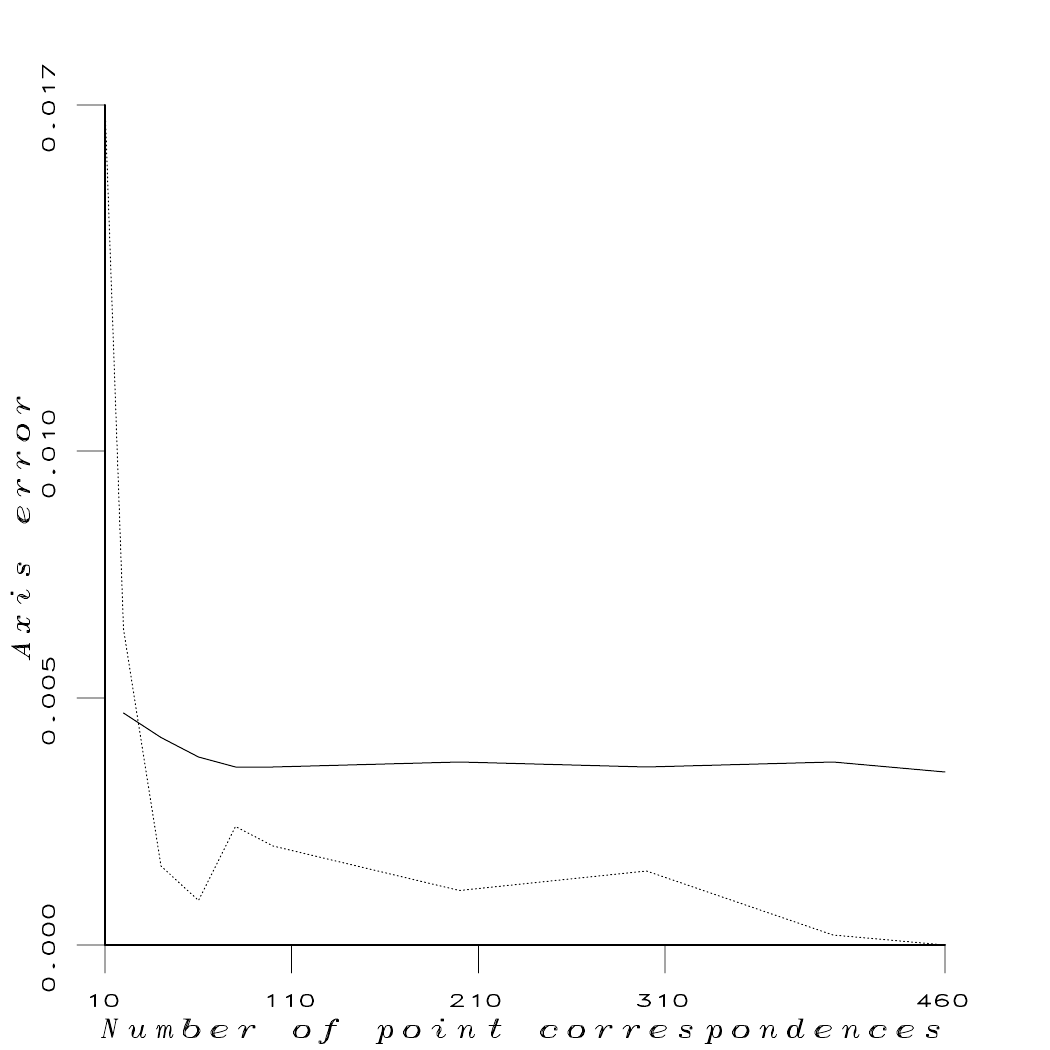}
\includegraphics[width=0.49\textwidth]{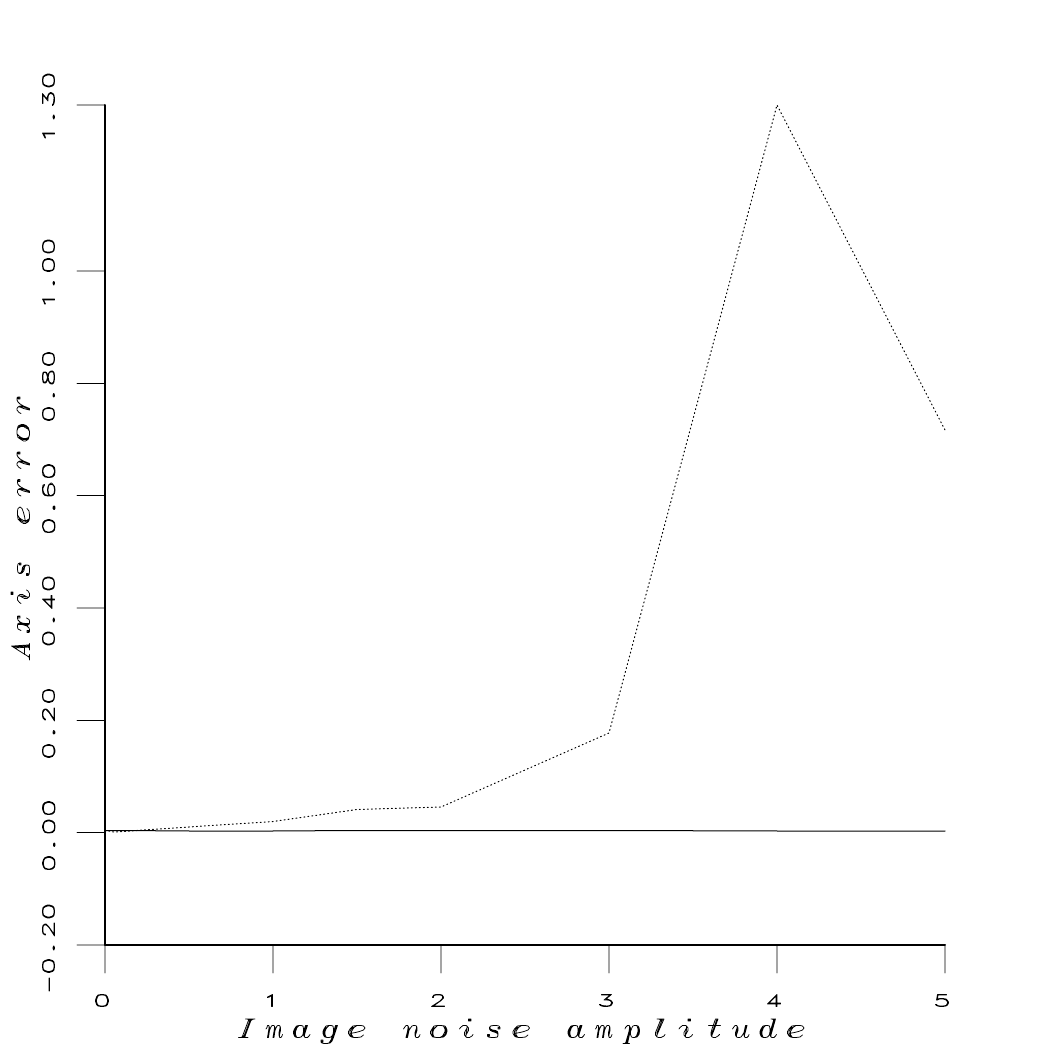}
\caption{Axis of rotation: The error between the computed direction of rotation and the ``true" one, as
a function of the number of correspondences (left) and as a function of image noise (right). The dashed
curves correspond to the Faugeras-Toscani linear method while the full curves correspond to the
rotation-then-translation implementation of our method.}
\label{fig:points-noise-axis}
\end{figure*}

\begin{figure*}[t!h!]
\centering
\includegraphics[width=0.49\textwidth]{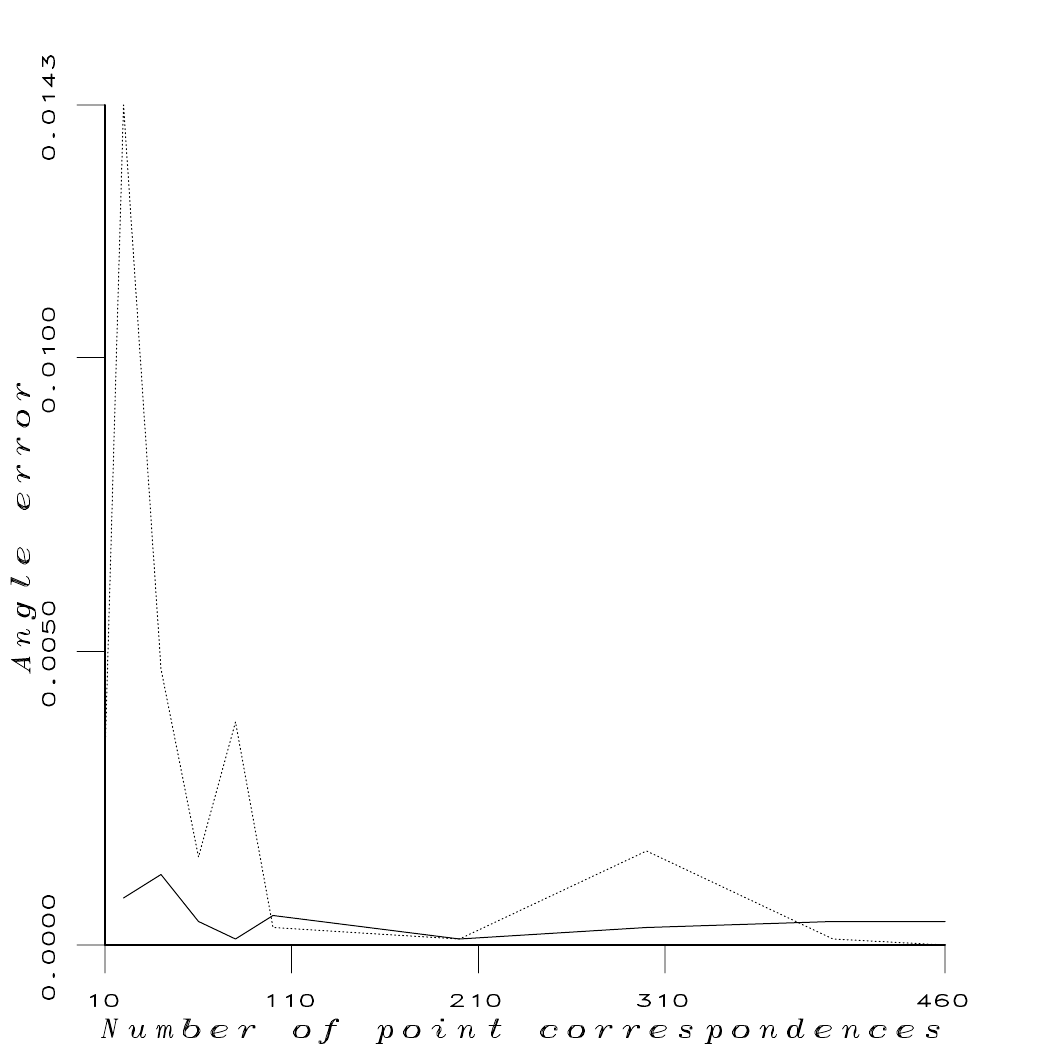}
\includegraphics[width=0.49\textwidth]{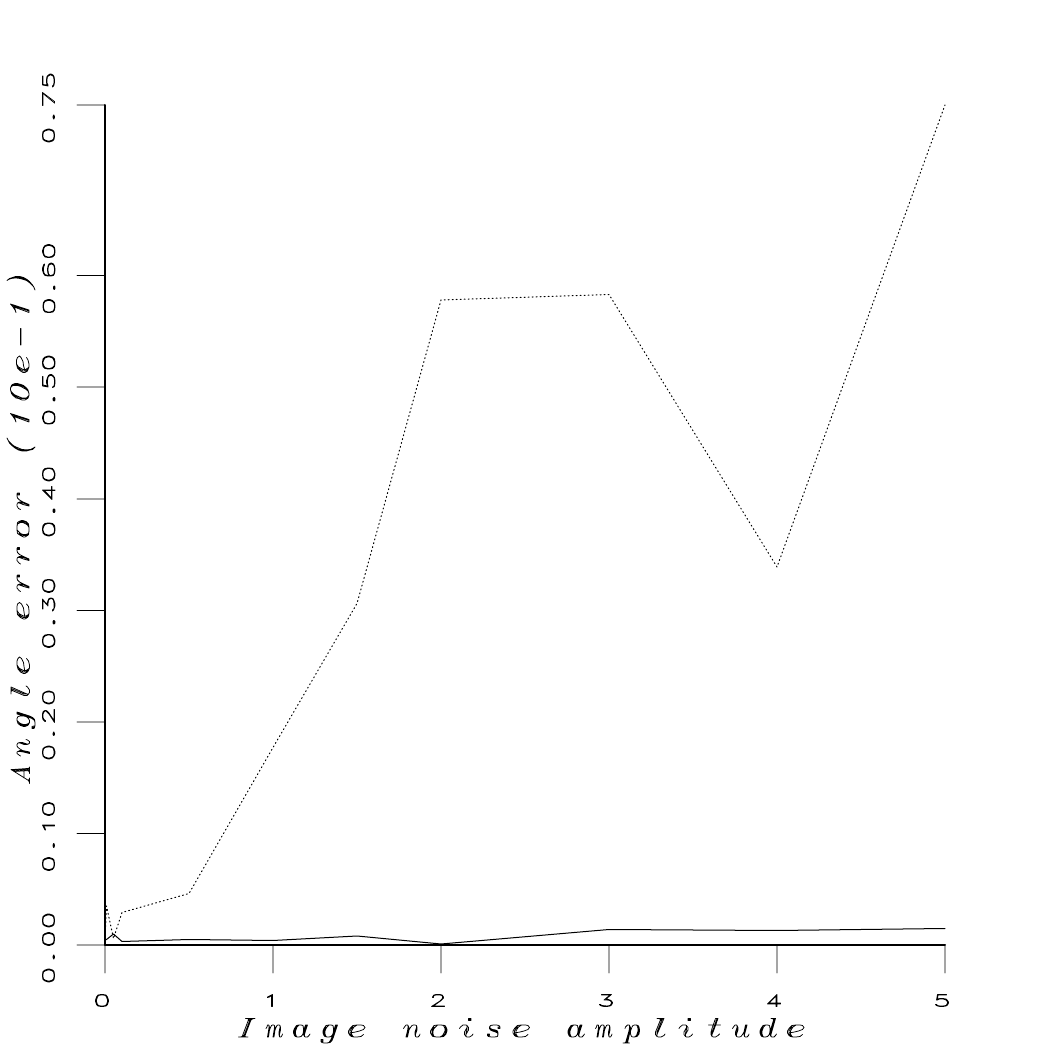}
\caption{Angle of rotation: The error between the computed angle of rotation and the ``true" one, 
measured in radians, as
a function of the number of correspondences (left) and as a function of image noise (right). The dashed
curves correspond to Faugeras-Toscani while the full curves correspond to the rotation-then-translation
implementation of our method.}
\label{fig:points-noise-angle}
\end{figure*}

\begin{figure*}
\centering
\includegraphics[width=0.49\textwidth]{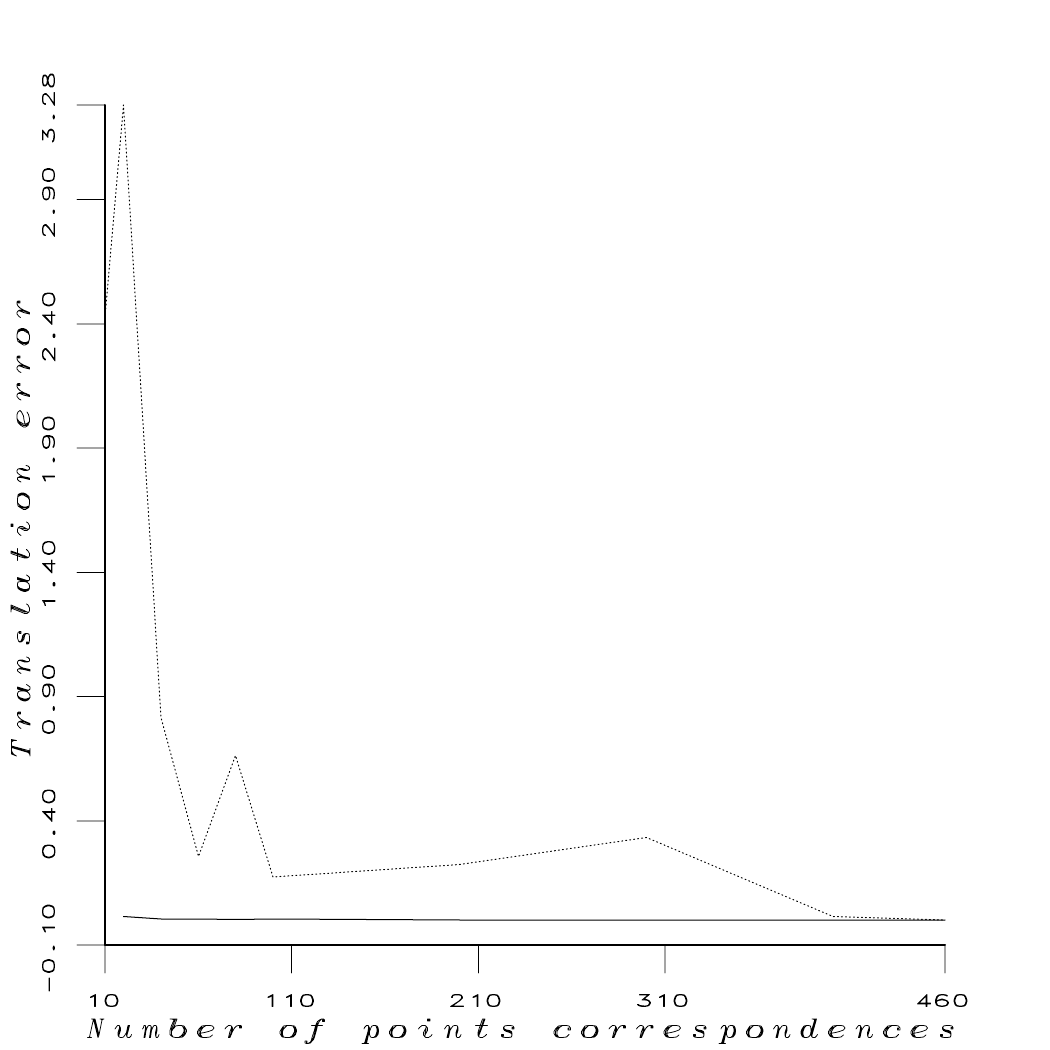}
\includegraphics[width=0.49\textwidth]{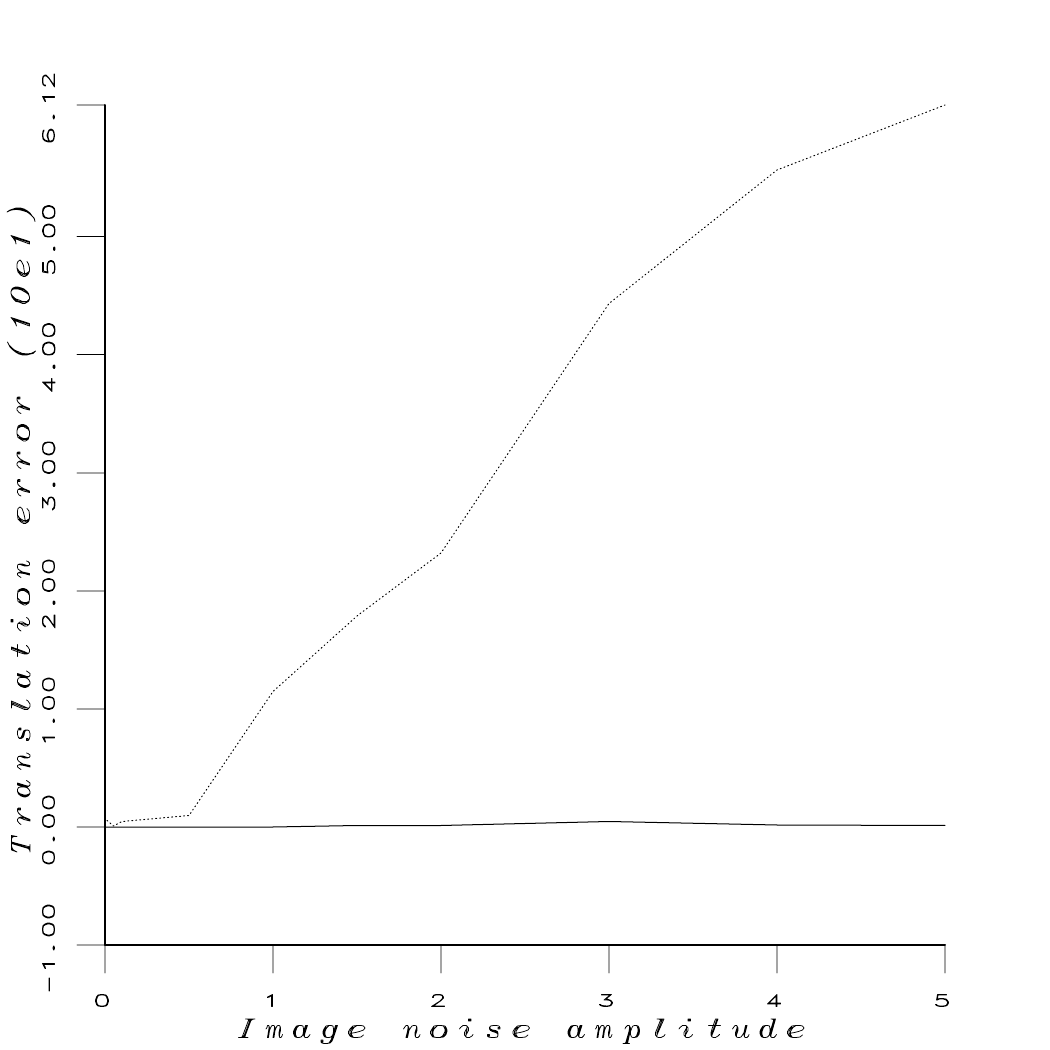}
\caption{Vector of translation: The error between the computed translation and the ``true" one, as
a function of the number of correspondences (left) and as a function of image noise (right). The dashed
curves correspond to Faugeras-Toscani while the full curves correspond to the rotation-then-translation
implementation of our method. The translation error is measured in centimeters. Notice the rapid
degradation of the estimation of translation in the presence of noise with the linear method.}
\label{fig:points-noise-transl}
\end{figure*}

A quantitative comparison between the Faugeras-Toscani method and the trust-region method is provided by
Figures~\ref{fig:points-noise-axis}, \ref{fig:points-noise-angle}, and \ref{fig:points-noise-transl}.
The dashed curves correspond to the Faugeras-Toscani method and the full curves correspond to the
rotation-then-translation implementation of our method. The errors in rotation (axis and angle) and
translation, as compared with the ground-truth solution, are shown as functions of the number of
correspondences (left) and as functions of increasing image noise (right). These curves correspond to
data otherwise shown in Tables~\ref{table:faugeras-points}, \ref{table:faugeras-noise},
\ref{table:lines-RthenT}, and \ref{table:lines-RthenT-noise}.

A second measure for the quality of a solution is estimated in image space. Independently of the
number of points (or lines) that are actually used for obtaining a solution, {\em all} 460 3-D points are
projected onto the image using that solution's parameters. Then, we compute the error (the Euclidean
distance) between the actual image locations and the projections of the 3-D points. More precisely
the error associated with a correspondence $i$ is:
\[
e_{i} = \left( (u_{i} - u_{i}^{s})^{2} + (v_{i} - v_{i}^{s})^{2} \right) ^{1/2}
\]
where $u_{i}$ and $v_{i}$ are the image coordinates of a point $i$ and $u_{i}^{s}$ and
$v_{i}^{s}$ are given by:
\begin{eqnarray*}
u_{i}^{s} & = & \alpha _{u} \frac{r_{11}^{s} X_{i} + r_{12}^{s} Y_{i} + r_{13}^{s} Z_{i} + t_{1}^{s}}{r_{31}^{s} X_{i} + r_{32}^{s} Y_{i} + r_{33}^{s} Z_{i} + t_{3}^{s}} + u_{0} \\
v_{i}^{s} & = & \alpha _{v} \frac{r_{21}^{s} X_{i} + r_{22}^{s} Y_{i} + r_{23}^{s} Z_{i} + t_{2}^{s}}{r_{31}^{s} X_{i} + r_{32}^{s} Y_{i} + r_{33}^{s} Z_{i} + t_{3}^{s}} + v_{0}
\end{eqnarray*}
Then, for each solution we estimate:
\begin{itemize}
\item the largest error;
\item the smallest error;
\item the average error;
\end{itemize}
over the 460 point correspondences. The values of these errors are reported in Tables \ref{table:faugeras-points} through \ref{table:points-RandT-noise}.

\subsection{Matching errors}

\begin{figure*}[htb]
\centering
\includegraphics[width=10cm]{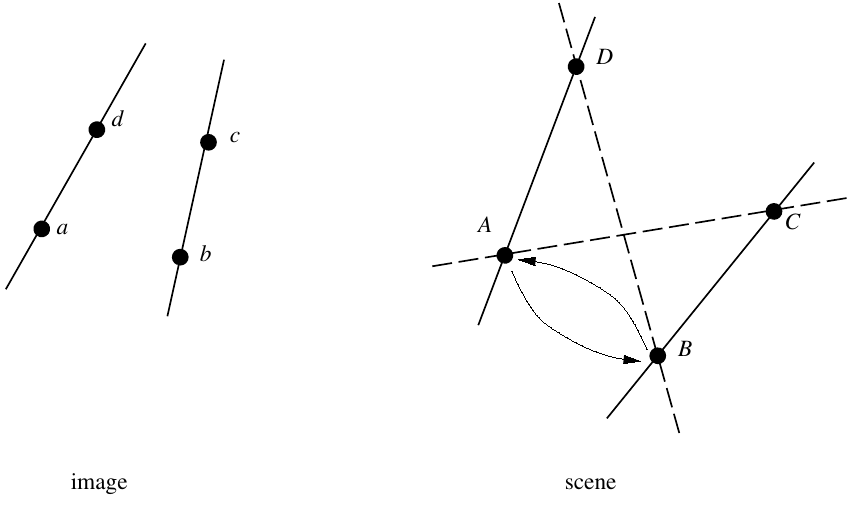}
\caption{From 4 correct point correspondences, one may define one line mismatched pair.}
\label{fig:mismatches}
\end{figure*}

In order to test the robustness of both the Faugeras-Toscani method and our method with respect to matching
errors, we studied the error between the ground-truth solution and solutions obtained with a variable
number of mismatches. A {\em mismatch}, or a matching error is defined as follows. Let $(A, B, C, D)$ be
4 scene points and let $(a, b, c, d)$ be 4 image points. Assume that the following 4 point correspondences
are correct: $(A-a, B-b, C-c, D-d)$. Moreover, one may associate 2 line correspondences with these 4 point
correspondences, $(AD-ad, BC-bc)$, e.g., Figure~\ref{fig:mismatches}. If $A$ is matched to $b$ and $B$ is
matched to $a$ we obtain a {\em point mismatched pair}, that is $(A-b, B-a)$, and a {\em line mismatched
pair}, that is $(BD-ad, AC-bc)$.

\begin{figure*}[h]
\centering
\includegraphics[width=0.49\textwidth]{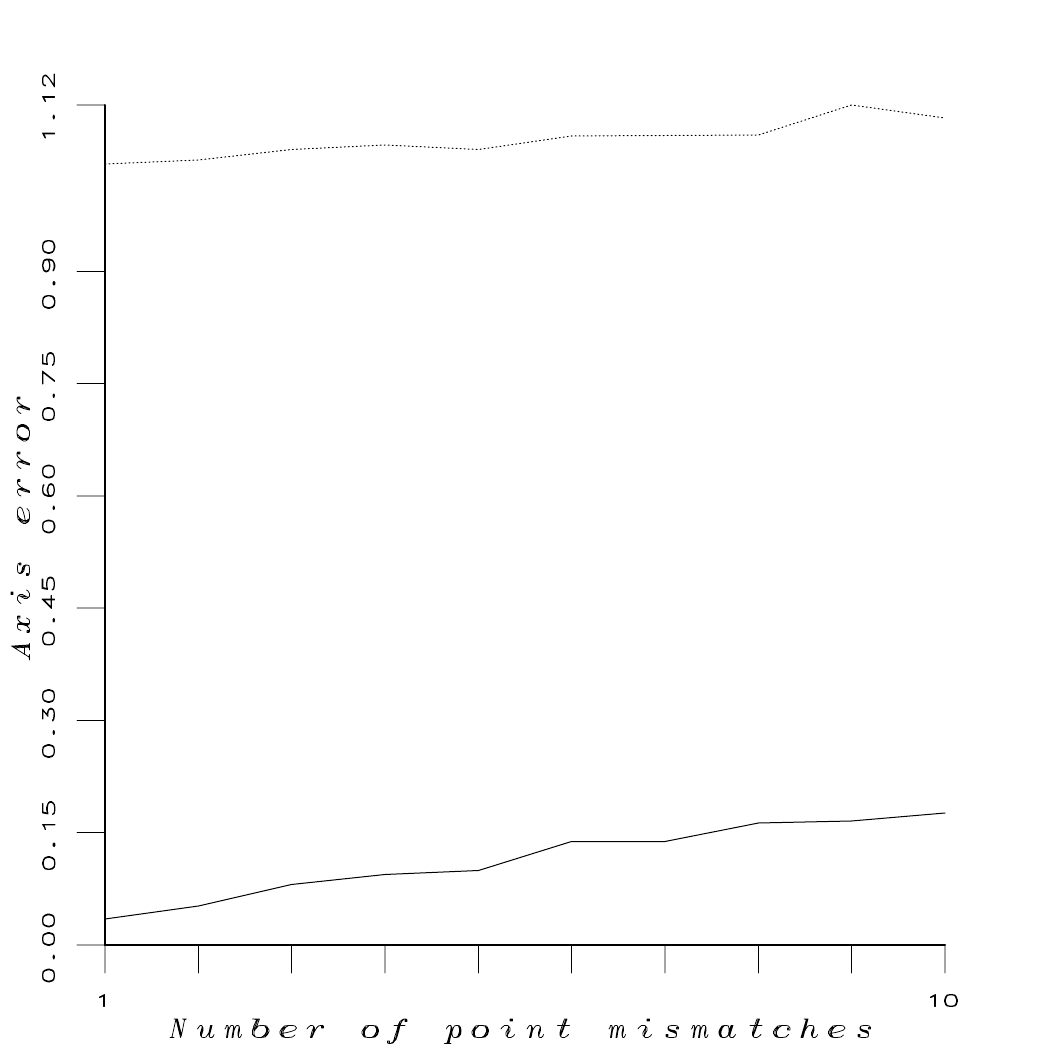}
\includegraphics[width=0.49\textwidth]{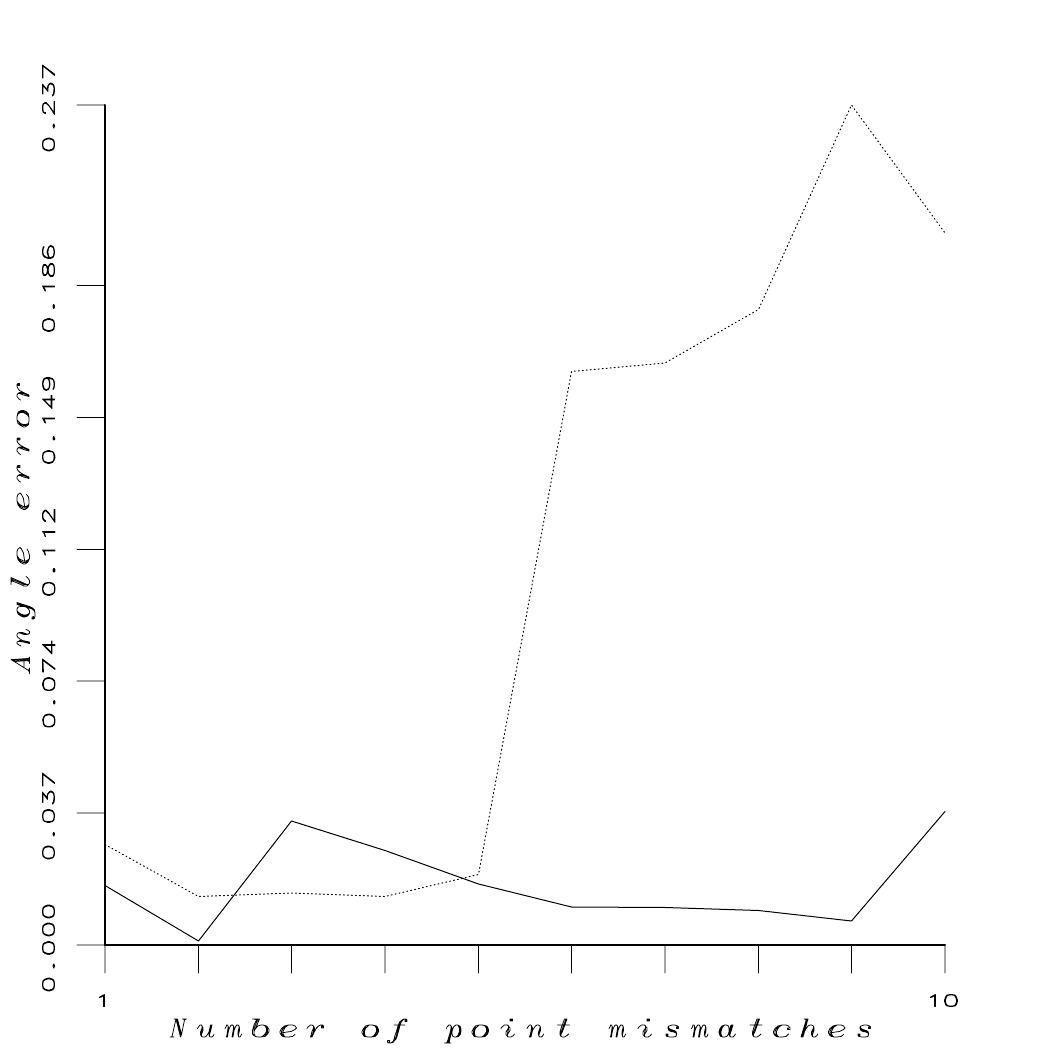}
\includegraphics[width=0.49\textwidth]{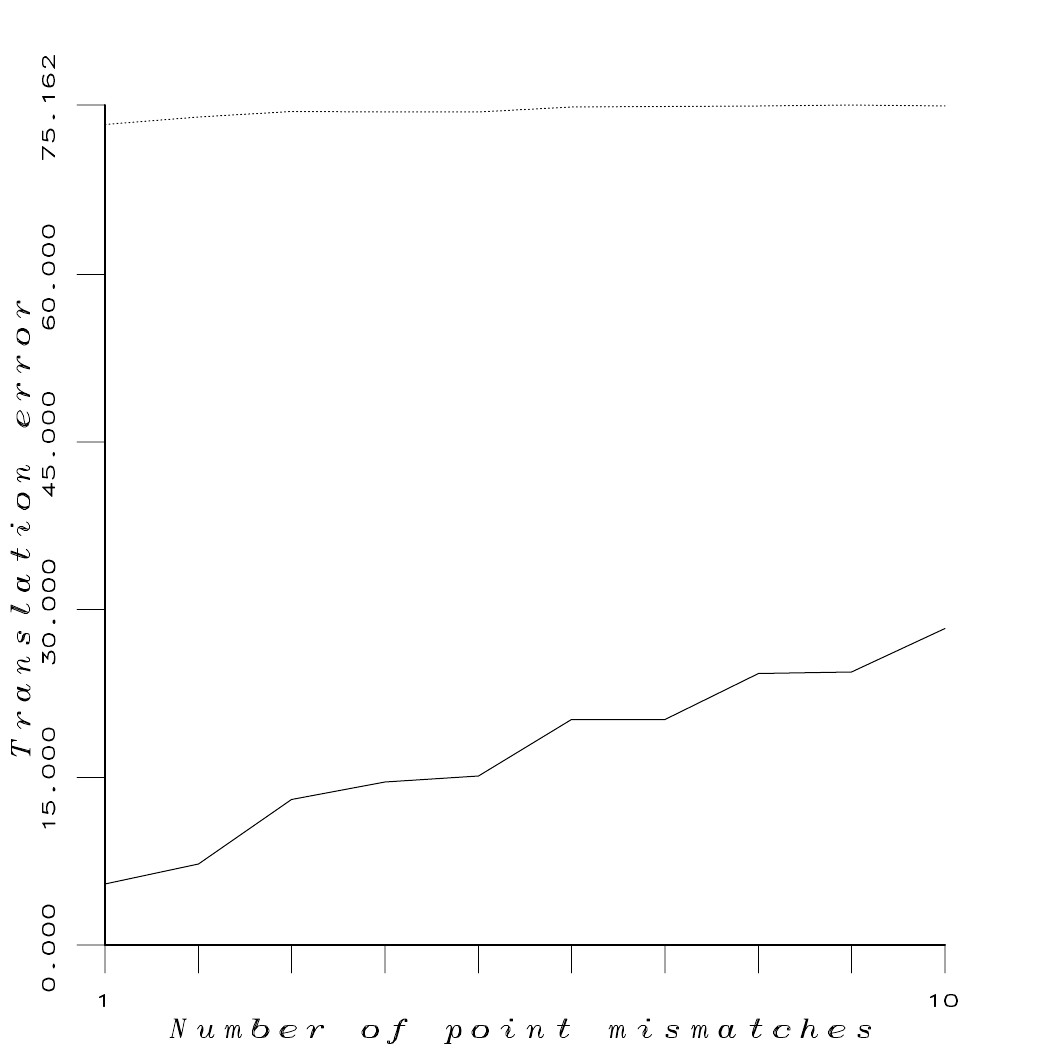}
\caption{Matching errors: The errors in rotation axis (top), rotation angle (middle), and translation
(bottom) as functions of the number of point mismatched pairs. The results were obtained with an initial set
of 400 points among which 1, 2, ... 10 point pairs were chosen at random and mismatched. The dashed
curves correspond to the Faugeras-Toscani method while the full curves correspond to the
rotation-then-translation implementation of the non linear method.}
\label{fig:matching-errors}
\end{figure*}

Figure \ref{fig:matching-errors} shows the errors in rotation (axis and angle) and translation as functions
of the number of the {\em point mismatched pairs}. These experiments show that the results obtained with the
non-linear method (full curves) degrade gently as the number of matching errors increases while the linear
method (dashed curves) give erroneous results.

\section{Discussion}

The method that we presented in this paper for estimating the exterior parameters of a
camera from line and/or point correspondences may be evaluated with respect to the following items:
\begin{itemize}
\item {\em Initialisation} -- the final result is independent of the initialisation. This is a dramatic improvement
with respect to other approaches using Newton's method, as reported in
\cite{Yuan89}, \cite{Lowe87}, and \cite{Lowe91}.
\item {\em Number of correspondences} -- The non linear method is more stable than the linear method when the
number of correspondences varies. This is particularly true for the estimation of the translation vector, Tables \ref{table:faugeras-points}, 
\ref{table:lines-RthenT}, \ref{table:lines-RandT}, 
and \ref{table:points-RandT}. 
\item {\em Accuracy} -- the algorithm nicely resists when noise is injected in the image. This is probably
one of the major advantages of the non linear method presented here with respect to linear methods, Tables \ref{table:faugeras-noise}, 
\ref{table:lines-RthenT-noise}, \ref{table:lines-RandT-noise}, 
and \ref{table:points-RandT-noise}.
\item {\em Efficiency} -- the rotation then translation implementation is more efficient
than the rotation and translation implementations. This is due to the fact that the trust region
algorithm minimizes over 4 parameters in the former case and over 8 parameters in the latter
case. In fact there is a compromise between efficiency and accuracy. One may be interested
in a fast algorithm which will provide a less accurate result. Ideally, with 30 line
correspondences, the rotation then translation algorithm converges in 0.03 seconds on a Sun/Sparc10, e.g., Table~\ref{table:lines-RthenT}. 
\item {\em Matching errors} -- when matching errors are artificially introduced in the data, as explained at the end of the previous section, 
the results obtained with the non linear method gently degrade as the number
of mismatches increases, while the linear method gives bad results whichever
the number of mismatches is, e.g., Figure~\ref{fig:matching-errors}.
\end{itemize}

To conclude, we believe that the pose determination method that we presented in this paper has those properties
that make it suitable to be used whenever robustness and accuracy are needed.
We also believe that the trust-region method that we presented in Section~4 and Appendix~A could beneficially be used to solve for other
non-linear minimization problems (not necessarily quadratic) in computer vision such as 
structure-from-motion, reconstruction from multiple views, or the estimation of the
fundamental matrix \cite{FaugerasLuongMaybank92}.

\appendix
\section{Minimization of the local quadratic form}

Consider the problem of minimizing a quadratic form
inside a sphere:
\begin{equation}
\min _{d} \{ \frac{1}{2}d^{T}Hd +g^{T}d : \|d\| \leq \delta \}
\label{eq:LQP}
\end{equation}
where $g\in I\!\!R^n$ , $H$ is a $n \times n $ symmetric matrix and $\delta$ is a positive number. Since the
constraint set is a compact, the problem has a solution. All existing methods for solving this
problem are based on the following theorem due to Gay \cite{kn:Gay} and to Mor\'{e} 
\& Sorensen \cite{kn:MoSo79}:
\begin{theor}  $d^{*}$ is a solution to (\ref{eq:LQP}) if and only if there exists $\mu_*\geq 0$ such that:
\begin{description}
 \item {(i)} $H+\mu_* I$ is positive semidefinite,
 \item {(ii)} $(H+\mu_* I)d^{*}=-g$,
 \item {(iii)} $\|d^{*}\|\leq \delta$ and $\mu_*(\|d^{*}\|-\delta)=0$.
\end{description}
Such a $\mu_*$ is unique.
\end{theor}

If $\mu$ can be estimated, the optimal solution to our minimization problem, can be easily
computed. In fact, the minimization of (\ref{eq:LQP}) represents an interesting case of non convex
optimization and in what follows we will point out a  complete characterization of a solution.
The solution itself is given by an algorithm which will be  outlined.

From  (iii) (Theorem 1) we see that, for $\mu_* > 0$, (\ref{eq:LQP}) has a solution on the boundary of its constraint set, i.e,
\[ \|d^{*}\|=\delta \]
and the problem is reduced to the problem of finding $\mu_*$ such that:
\begin{equation}
\label{eq:phi1}
         \phi (\mu_*) = \|d(\mu_*)\|=\delta
\end{equation}
where $d(\mu)$ is a solution of $(H+\mu I)d=-g$.

 Let $\lambda_{1} \geq \lambda_{2} \geq \ldots \geq \lambda_{n}$ be the eigenvalues of $H$ (a symmetric matrix) and let $u_{1}, u_{2}, \ldots, u_{n}$ be their corresponding eigenvectors. The function $\phi$ can be made explicit using the
eigensystem of $H$:
\begin{equation}
\label{eq:phi2}
\phi(\mu)= \left (\sum_{i=1}^{n} \frac{(u ^{T}_{i}g)^{2}}{(\lambda _{i}+\mu)^{2}}\right)^{1/2}
\end{equation}
It is easy to verify the following properties of the function $\phi$:
\begin{enumerate}
\item $\phi(\mu)$ is positive, strictly decreasing in $]-\lambda_{n},+\infty[$ and
 $\phi(\mu) \rightarrow 0$ as $\mu \rightarrow \infty$.
\item $\phi^{2}(\mu)$ is rational in $\mu$ with second-order poles on a subset of the negatives of the eigenvalues of $H$ (see Hebden \cite{kn:Heb})
\end{enumerate}

Thus, when $\mu_* > 0$ (e.g., when the Gauss-Newton approximation of the Hessian is being used)  there exists an unique solution $d^*=d(\mu_*)$ which is a solution of eq. (\ref{eq:phi1}). Such a couple $(\mu_*,d^*)$  can be determined by a very efficient  method developed by Hebden \cite{kn:Heb}. Given a $\mu_o \geq 0$ such that $\phi(\mu_o)>\delta$, Hebden proposed to update $\mu_{l+1}$ from
$\mu_{l}$ as following:
 \begin{equation}
   \label{eq:heb}
   \mu_{l+1}=\mu_{l}+\frac{\phi(\mu_{l})}{\phi'(\mu_{l})}\frac{(\delta -\phi(\mu_{l})
)}{\delta}.
\end{equation}
In fact, Hebden's algorithm can be viewed as Newton's method for the zero-finding problem:
\begin{equation}
   \label{eq:new}
   \psi(\mu)=\frac{1}{\delta}- \frac{1}{\phi(\mu)}=0, \;\; \mbox{for}\; \mu \in ]-\lambda_{n}, +\infty[
\end{equation}
The most important feature of (\ref{eq:heb}) is that usually the number of iterations
 required to produce an acceptable approximation of solution  $\mu_{*}$ is very small
 since $\psi$ is convex, almost linear, and strictly decreasing on $]-\lambda_{n}, +\infty[$ . 
Moreover, the computation of the Cholesky factorization of $H+\mu I$ makes it
 possible to implement (\ref{eq:heb}) without explicit knowledge of the eigensystem of 
$H$ whenever $\mu \in ]-\lambda_{n}, +\infty[$. More precisely, it is easy to see that
\[ \phi(\mu)=\|d(\mu)\|, \hspace{1cm} \phi'(\mu)=-[1/\phi(\mu)]d(\mu)^{T}(H+\mu I)^{-
1} d(\mu), \]
where $(H+\mu I)d(\mu)=-g$. So, if $R^{T}R$ is the Cholesky factorization of $H+\mu_{l} I$ 
with $R$ being an upper triangular matrix then we have
\[
       \frac{\phi(\mu_{l})}{\phi'(\mu_{l})}=- \frac{\phi^{2}(\mu_{l})}{d(\mu_{l})^{T}
(R^{T}R)^{-1}d(\mu_{l})}=- \frac{\phi^{2}(\mu_{l})}{\|(R^{T})^{-1}d(\mu_{l})\|^{2}}
\]

Therefore, Hebden's algorithm which finds the minimum of the local quadratic form can be
summarized as follows:
\begin{itemize}
 \item {\em Initialization}: Let $\mu_{o}\geq 0$  and $\phi(\mu_{o})>\delta$.
 \item {\em Iteration l=0,1,\ldots }
  \begin{description}
  \item {l.1} Factor $(H+\mu_{l} I)=R^{T}R$.
  \item {l.2} Solve $R^{T}Rp=-g$.
  \item {l.3} If $|\|p\|-\delta|<\epsilon $ then stop: $\mu_{l}$ is an approximation
 of $\mu^{*}$ and $p$ is a solution of (\ref{eq:LQP}). 
  \item {l.4} Otherwise, solve $R^{T}q=p$.
  \item {l.5} Set
\begin{equation}
   \label{eq:heb1}
   \mu_{l+1}=\mu_{l}+ \left(\frac{\|p\|}{\|q\|}\right)^{2}\frac{(\|p\|-\delta)}{\delta}.
\end{equation}
  \item {l.6} Set l:=l+1 and return to l.1.
 \end{description}
\end{itemize}

In practice, this algorithm converges in average after 4 iterations.

\section*{Acknowledgments} The authors acknowledge Claude Inglebert, Homer Chen, Roger Mohr, Long Quan, Fadi Dornaika, and Thomas Skordas for
their insightful comments.
\balance
\bibliographystyle{plain}

\newpage

\end{document}